\documentclass{article}

\PassOptionsToPackage{numbers, compress}{natbib}
\usepackage{iclr2025_conference,times}
\iclrfinalcopy

\usepackage[utf8]{inputenc} 
\usepackage[T1]{fontenc}    
\usepackage{url}            
\usepackage{booktabs}       
\usepackage{amsfonts}       
\usepackage{nicefrac}       
\usepackage{microtype}      
\usepackage{xcolor}         
\usepackage{pdflscape}
\usepackage{amsmath,amsfonts}
\usepackage[leftcaption]{sidecap}
\usepackage{graphicx}
\usepackage[colorlinks=true, allcolors=blue]{hyperref}
\usepackage{subcaption}
\usepackage{array}
\usepackage{appendix}
\usepackage{rotating}
\usepackage{enumitem}
\usepackage{natbib}
\usepackage{multirow}

\title{Do LLMs have Consistent Values?}

\newcommand{\amir}[1]{{\color{green}Amir: #1}}
\newcommand{\ella}[1]{{\color{blue}Ella: #1}}
\newcommand{\naama}[1]{{\color{magenta}Naama: #1}}

\newcommand{\comment}[1]{}
\newcommand{\secref}[1]{Section \ref{#1}}
\newcommand{\figref}[1]{Figure \ref{#1}}
\newcommand{\tabref}[1]{Table \ref{#1}}

\author{
  \textbf{Naama Rozen}\\
  Tel-Aviv University\\
  \texttt{naamarozen240@gmail.com}\\
  \And
  \textbf{Liat Bezalel} \\
  Tel-Aviv University\\
  \texttt{liatbezalel@mail.tau.ac.il}\\
  \And
  \textbf{Gal Elidan} \\
  Google\\
  Hebrew University\\
  \texttt{elidan@google.com}\\
  \And
  \textbf{Amir Globerson} \\
  Google\\
  Tel-Aviv University \\ 
  \texttt{amirg@google.com}\\
  \And
  \textbf{Ella Daniel} \\
  Tel-Aviv University\\
  \texttt{della@tauex.tau.ac.il}\\
}
\date{}
\begin{document}
\maketitle

\begin{abstract}
 Large Language Models (LLM) technology is constantly improving towards human-like dialogue. Values are a basic driving force underlying human behavior, but little research has been done to study the values exhibited in text generated by LLMs. Here we study this question by turning to the rich literature on value structure in psychology. We ask whether LLMs exhibit the same value structure that has been demonstrated in humans, including the ranking of values, and correlation between values. We show that the results of this analysis depend on how the LLM is prompted, and that under a particular prompting strategy (referred to as ``Value Anchoring'') the agreement with human data is quite compelling. Our results serve both to improve our understanding of values in LLMs, as well as introduce novel methods for assessing consistency in LLM responses.
\end{abstract}

\section{Introduction}
A key goal of Large Language Models (LLMs) is to produce agents that will be able to communicate in a ``human-like'' fashion. However,  human communication is characterised by some level of consistency within an individual, as well as variability between individuals. This raises a key question: during a single conversation with an LLM, does the ``LLM-persona'' resemble a single human? Furthermore, across multiple conversations, can LLMs produce multiple personas that resemble a population of humans? If this is indeed possible, how can such personas be elicited to best resemble psychological characteristics observed in human populations?

This question has only recently begun to be addressed. For example \cite{aher2023using} show how probing LLMs with different names leads to variability which in some cases agrees with that of human populations. Here our focus is on understanding whether an LLM in a single conversation can exhibit a psychological characteristic profile that is similar to that of humans. This is a highly challenging question, since it requires analyzing a complete conversation and evaluating whether it conceivably could have been generated by a single individual.

In order to stand on more quantitative ground to evaluating the quality of output relative to human research, we turn to the well established field of value psychology. Namely, we aim to quantify the values that LLM responses are aligned with, and whether these are in agreement with the value hierarchy and structure observed in humans. The question of values in LLMs has rarely been studied, and is of naturally  broad interest. As an example of recent work, \cite{fischer2023what} prompt an LLM with a description of a profile of an individual characterized by a value and check whether generated text is consistent with this description. Our focus is very different, and asks whether an LLM response is in agreement with what we expect human responses to look like given research in the field.

Values are basic motivations that play a foundational role in psychology, influencing perceptions and behaviors across various domains \citep{sagiv2022personal, sagiv2017personal}, and representing fundamental aspects of human personality \citep{roberts2022personality}. Research has consistently demonstrated their enduring influence over behavior across time and contexts \citep{sagiv2017personal}. 

\begin{SCfigure}
    \centering
    \includegraphics[trim=0 0 0 0,clip,width=0.5\linewidth]{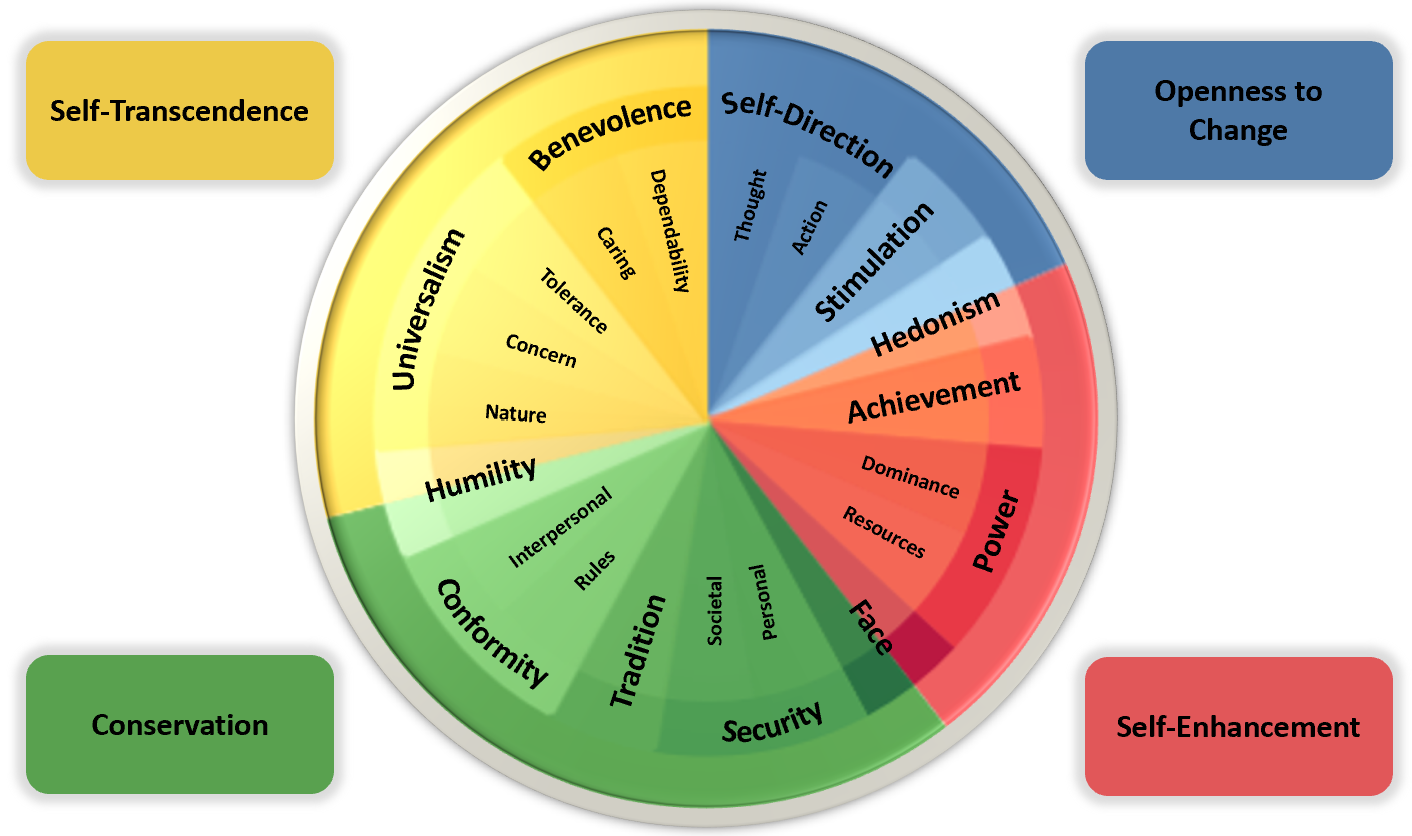}
    \caption{Circular motivational continuum of 19 values in the refined value theory. Source: \cite{schwartz2012refining}. A value aligns with values that are adjacent on the circle and conflicts with those opposite to it. For example, self direction aligns with stimulation, and both conflict with conformity.} 
    \label{fig:Theorized circle}
\end{SCfigure}

One prominent framework for studying values, the Theory of Basic Human Values \citep{schwartz1992universals}, outlines 19 core values, categorized by motivational goals \citep{schwartz2012overview}. These values can be simplified into a two-dimensional structure: conservation vs. openness to change, and self-enhancement vs. self-transcendence. The theory describes interrelations among values, suggesting that motivations driving some values are compatible with those driving other values, yet conflict with those underlying yet others. For instance, pursuing independence and creativity (self-direction) aligns with seeking change and variability (stimulation), but conflicts with an emphasis on the status quo (conformity). See \figref{fig:Theorized circle} for the theorized circle. One of the key aspects of the theory is its cross-cultural coverage: it was developed to apply across populations, and tested in nearly 100 countries across all continents of the world, identifying points of commonality and differences among these populations \citep{sagiv2022personal}. The use of the theory allows a stable and extremely general baseline of human values to compare LLMs.  
Hundreds of samples demonstrate individual differences in value importance. Importantly, they also demonstrate a universal hierarchy, where people are more likely to stress some values over others, notwithstanding the existing variability. For example caring for close others ranks high, while values related to dominance hold less importance across societies \citep{schwartz2001value}. There is also ample empirical evidence that compatible values tend to be correlated in humans \citep{skimina2021between, daniel2023development, schwartz2022measuring}, and are thus a ``marker'' of a human-like value system. To summarize the above, human data pertains to both first-order statistics of values (i.e., which values rank high or low across the population), and second-order statistics (i.e., how are different values within an individual correlated). 

\sidecaptionvpos{figure}{c}

Our key question is therefore whether LLM responses demonstrate the same statistical behavior observed in humans with respect to both value-ranking and value-correlations. Note that the question of value-correlations is of particular interest, because it allows bench-marking  the extent to which responses of an LLM demonstrate a coherent ``persona''. For example, while it is possible for a person to give a high score to Power Dominance, that person is unlikely to give a high score to Benevolence, since these are contradicting values.


To study this question quantitatively, we present LLMs with a value questionnaire (the Portrait Value Questionnaire—Revised -- PVQ-RR-- from  \cite{schwartz2017refined}, a well-established measure of values), and prompt them to answer all the questions in a single session (i.e., in the same context window). We then analyze the provided answers, putting specific emphasis on the correlation between answers in the same session. 

We analyze two recent LLMs: GPT 4 and Gemini Pro, as well as four open models: Llama 3.1 8B, Llama 3.1 70B, Gemma 2 9B, and Gemma 2 27B.\footnote{Our analysis also included GPT-3.5 and Palm2, which produced qualitatively similar results.} Our results show that standard prompting of LLMs {\em does not} result in a population of human-like personas. We go on to explore prompting the LLMs with other prompts that provide additional information about the LLM persona. In particular we consider names \citep{aher2023using} and persona descriptions. In addition, we consider a novel prompt which we refer to as a ``Value Anchor'', which instructs the language model to answer as a person emphasizing a given value. We find that with these prompts, and in particular with the Value Anchor prompt, the overall first and second order statistics of the LLM responses
 closely mirror those of human subjects. Perhaps most surprising is our finding that the correlation between values agrees with the well known Schwartz circular model for correlations between values. We furthermore provide an explanation for how this correlation comes about.
\comment{
Our main findings are thus that LLMs cannot be considered as having a consistent, human like, value system that can be studied as representative of the values of the LLM. However, with appropriate prompting, LLMs can exhibit various coherent value personas similar to that of a population of humans. In other words, they answer a set of questions in a way that is consistent with a response from multiple human subjects. 
}We provide information for best prompts and settings for this aim. In addition, we include six datasets comprising 300 personas each, generated by the models. In conclusion, our results demonstrate the utility of using psychological theory for evaluating consistency of personas generated by LLMs.

\section{Related Work}
{\bf Values in LLMs:}
Our work is based on the Schwartz theory of Personal Values, a highly accepted theory within personality psychology \citep{sagiv2022personal}. Values are abstract goals, defining the end states individuals aspire for (e.g., safety, independence), used to direct judgements and behaviors \citep{schwartz1992universals,schwartz2012overview}. Individuals typically prioritize their values, so that values stemming from compatible motivations are similarly important, while values stemming from conflicting motivations are prioritized differently. These associations were replicated across hundreds of samples, across the world \citep{pakizeh2007basic,skimina2021traits}, and make value theory especially useful to identify the coherence of the value profiles created by LLMs. 
Several studies assumed LLMs can be characterized as operating on the basis of a single set of values, taking an ``LLMs as individuals'' approach. \cite{fischer2023what} tested whether ChatGPT could comprehend human values by providing it with value-related prompts and analyzing whether its responses matched the intended value category. 
A second study \cite{lindahl2023unveiling}, compared ChatGPT's values to those observed in the World Value Survey, while another \cite{miotto2022gpt} investigated how temperature influences GPT-3's responses to the Human Value Scale. Scherrer and colleagues \cite{ScherrerSFB23} studied responses of LLMs prompts evaluating moral positions, especially in ambiguous settings. A recent study by \cite{hadar2024assessing}, tested the value-like constructs embedded in LLMs and revealed both similarities and differences between LLMs and humans' values. 
\cite{kovavc2023large} challenged those studies by establishing that context starkly influences values expressed by ChatGPT. They found significant variability in ChatGPT's value expression in response to contextual changes, threatening the notion of stable characteristics of LLMs. Building upon the insights in \cite{kovavc2023large}, our study posits that upon providing controlled variability of context, LLMs can elicit a population of multiple personas. In this regard, we aim to further explore the accuracy of LLMs' mimicking abilities within a controlled experimental framework. 
\smallskip\newline
{\bf Prompting LLMs:}
There is extensive research on prompt design for mimicking individual characteristics in LLMs \citep{liu2023pre}. Approaches use specific scenarios \citep{hadar2023plasticity}, questionnaire items \citep{jiang2023personallm}, simulation of social identities or areas of expertise \citep{salewski2024context}, utilization of titles and surnames representing genders and ethnicities \citep{aher2023using}, and other demographic information \citep{argyle2023out}. Additionally, researchers explored the use of designated personas \citep{safdari2023personality}, and employed RLHF \citep{li2023theory} to guide LLMs to reflect distinct personality traits. Despite this extensive body of work, to our knowledge, no study has directly compared the various prompting techniques to determine which approach yields responses that simulate within-session psychological characteristics of an individual best.
\smallskip\newline
{\bf Temperature in LLMs:}
Adjusting the temperature stands as a common practice for introducing variability in LLM responses \citep{miotto2022gpt}. However, consensus is lacking on the optimal temperature setting in simulating psychological characteristics. Existing research includes use of mostly two temprature settings: 0.7 and 0. Some researchers advocate for higher temperatures to boost creativity \citep{salewski2024context}, yet this can also introduce more noise into the data \citep{gunel2020supervised}. Conversely, setting the temperature to zero minimizes variability, enhancing replicability \citep{li2023theory}, albeit posing challenges for variance-dependent analysis \citep{hagendorff2023human}. 
Our framework enables us to explore how temperature adjustments impact the ability of LLMs to simulate human characteristics across multiple datasets.
\smallskip\newline
{\bf Evaluating the Quality of Persona Generation in LLMs:}
The ability of LLMs to mimic and portray human characteristics is a focus of intense research \citep{binz2023using, ouyang2022training}. LLMs can express psychological traits and attributes similar to human individuals \citep{li2023theory, stevenson2022putting}, and even simulate diverse populations \citep{deshpande2023toxicity, salewski2024context}. However, we are only beginning to understand the coherence of
these LLM-generated characteristics in mirroring human psychological profiles \citep{aher2023using, kovavc2023large}, and how to reliably produce such responses. We are specifically challenged to evaluate the coherence of the resulting psychological profiles. The literature suggested a number of approaches, including an open-ended interview with LLM-generated personas in order to assess the consistency between their intended characters and the responses \citep{wang2024incharacter}. In addition, one may apply an additional ``judge'' LLM in order to check an LLM persona \citep{gupta2024bias}. Finally, \cite{jiang2023personallm} assessed coherence with a description used to prompt the LLM. 
Our study extends upon this line of research by applying well established characteristics of human psychology to investigate the quality of LLM generated personas.

\section{Method}
In this section, we introduce the experimental design, models and prompts. 
The code and data are provided as supplementary files in the submission.

{\bf The Value Questionnaire:}
Our key goal was to assess responses of LLMs to questionnaires used to measure 
values in human subjects.
Specifically, we considered the commonly used 57-item Portrait Value Questionnaire—Revised (PVQ-RR; \citep{schwartz2017refined}), developed to measure the 19 values in the Schwartz’s theory. The questionnaire describes fictional individuals and what is important to them. For example: ``It is important to him/her to take care of people he/she is close to'' (an item measuring benevolence-care values). For each such item, the subject is requested to indicate on a 6-point scale to what degree the persona they form is similar to the person described. Answers are categorical and range from a value of 1 (indicating ``not like me at all'') to 6 (indicating ``very much like me''). See Appendix for instructions and more example items from the questionnaire.
\smallskip \newline
{\bf Models Used:}
We employed six prominent LLMs, specifically OpenAI’s GPT-4, Google’s Gemini Pro, Llama 3.1 8B, Llama 3.1 70B, Gemma 2 9B, and Gemma 2 27B. Each model was prompted with the five prompts (see \secref{sec:prompts}), 300 times overall. Half of the runs applied the male-version of the questionnaire, and half the female version. The entire process was conducted twice, once with the temperature parameter set to $0.0$ and once with it set to $0.7$, resulting in the generation of 20 datasets for analysis.

\subsection{Prompts \label{sec:prompts}}
As mentioned above, we would like to measure the response of LLMs to PVQ-RR. However, as with 
many other LLM applications, the way the model is prompted has a significant effect on output.
The instructions of the PVQ questionnaire were similar to those in prior research, but with added text instructing the LLM not to elaborate. This resulted in the LLM producing only the value scores, thereby simplifying processing and analysis. LLMs were prompted to assess their likeness to the 57 descriptions incorporated in the PVQ-RR. Following each prompt, the LLM was provided with all 57 items of the questionnaire in one administration.\footnote {We also conducted a serial prompting analysis for the Llama models. The results comparing batch administration to serial administration are detailed in the Appendix \figref{fig:CorHeatmap_batchserial} and \tabref{tab:sum_square_scores_matrices_serial}} The study utilized a basic prompt, as well as four different prompts below that vary instructions to create multiple personas.

    {\bf Basic prompt:} This prompt mirrors the adapted instructions of the PVQ-RR questionnaire without additional modifications. The prompt is structured as follows: “\textit{For each of the following descriptions, please answer how much the person described is like you from 1 (Not like me at all) to 6 (Very much like me), without elaborating on your reasoning.}”.
    \smallskip\newline
    {\bf Value Anchor prompt:} This prompt adds an anchor of value importance using identification with an item used in an additional value questionnaire, akin to the approach outlined in the study by \cite{jiang2023personallm}. Participants are instructed as follows: ``\textit{For each of the following descriptions, please answer how much the person described is like you from 1 (Not like me at all) to 6 (Very much like me), without elaborating on your reasoning. Answer as a person that is [value]}''. Here ``[value]'' is taken from the Best-Worst Refined Values scale \citep{lee2019testing}. As a result, the prompts refer conceptually to the same values that are measured using the PVQ-RR, yet do not refer directly to the value items to be answered in response to the prompt. Examples of these anchor items include ``protecting the natural environment from destruction or pollution'' (universalism-nature) or ``obeying all rules and laws'' (conformity-rules). Please refer to \ref{sec:BWVr list} in the appendix for the complete list of anchor items.
\smallskip\newline
    {\bf Demographic prompt:} Drawing from the methodology of \cite{argyle2023out}, this prompt extends the original prompt by incorporating additional demographic details. LLMs are asked to provide ratings based on the following prompt: ``\textit{For each of the following descriptions, please rate how much the person described is like you, using a scale from 1 (Not like me at all) to 6 (Very much like me), without elaborating on your reasoning. Answer as a [age]-year-old who identifies as [gender], working in the field of [occupation], and enjoys [hobby].}'' The age, gender, occupation and hobby were randomly allocated for each prompt from a predefined list or range. The  age range specified was between 18 and 75, with gender options including male, female, non-binary, and other, adapted from the National Academies of Sciences, Engineering, and Medicine \citep{national2022measuring}. Occupations were sourced from the World Values Survey (WVS-7; \citep{haerpfer2022world}), while hobbies were chosen from established lists supplied by The Activity Card Sort (ACS-UK; \citep{laver2016face}). The lists of occupations and hobbies are available upon request.
\smallskip\newline
    {\bf Generated Persona prompt:} In line with the methodology of \cite{cheng2023marked}, we directed the models to craft personas. Our instruction was formulated as: ``\textit{Create a persona (2-3 sentences long):}'', with the temperature set at $0.7$ to stimulate the models' creativity. An example of a persona generated by Gemini Pro is as follows: ``Emily is a 25-year-old marketing manager who is passionate about her career and loves spending time with her friends and family. She is always looking for new ways to improve her skills and knowledge, and she is always up for a challenge.'' Using these generated personas, we subsequently prompted the model as follows: ``\textit{For each of the following descriptions, please rate how much the person described is like you, using a scale from 1 (Not like me at all) to 6 (Very much like me), without elaborating on your reasoning. Answer as: [persona].}'' 
\smallskip\newline
    {\bf Names prompt:} In line with a study by \cite{aher2023using}, the prompts comprised titles (i.e., Mr., Ms., and Mx.) followed by surnames representing five distinct ethnic groups. From the 500 names cataloged in the previous study, we randomly generated 300 unique combinations of titles and names, including 60 from each ethnic group. The prompt was structured as follows: ``\textit{For each of the following descriptions, please rate how much the person described is like you, using a scale from 1 (Not like me at all) to 6 (Very much like me), without elaborating on your reasoning. Answer as [title + name]}''. The complete list of titles and names are available upon request.

\subsection{Data Analysis}
In what follows we use the following notation. Let $V=19$ be the set of value types studied. Each question in the questionnaire pertains to a particular item within the set of values $i\in V$. Furthermore, for each value there are $R=3$ question variants. See \secref{sec:variants} in the Appendix for example variants.
Recall that the answer to each question is a number on a 6-point scale. For each LLM and prompt type, we presented the questionnaire $N$ times. The difference between each of these could be different personas, names, temperature sampling etc. 
Thus the overall set of answers corresponds to a set of values $X_{i,j,k}\in\{1,\ldots,6\}$ 
where $i=1,\ldots, V$ and $j=1,\ldots, R$, $k=1,\ldots, N$.

When comparing to human data, we used the study in \cite{schwartz2022measuring}. The data is from 49 cultural groups. The total number of participants was 53,472, the mean age was 34.2, (SD = 15.8), with 59\% females.  Their data is stored at the Open Science Framework and is available \href{https://osf.io/w9as3/?view_only=e1f02bf232c34d39b9884398b4f2df63}{here}.

\subsubsection{Value Rankings \label{sec:rankings}}
Although there is variability between individuals in their prioritization of values, there are values that tend to be ranked as more important than others across cultures and samples. Those suggest there are underlying principles that give rise to value hierarchies. The similarity in value importance across cultures is referred to as the universal value hierarchy \citep{schwartz2001value, schwartz2022measuring}. Our first question for analysis was whether this hierarchy is also reflected in LLM data. Namely, do LLMs tend to rank the same values as high or low as human subjects do. 

To obtain LLM rankings for a given set of LLM answers, we assigned a score $v_i$ to value $i$, where $v_i$ was the average score given to the three items measuring this value by the LLM (i.e. the average of $X_{i,\cdot,\cdot}$). From this score, we subtracted the average score given to all value items within the conversation, thus centering the data. Centring is the recommended practice in value research \citep{schwartz1992universals, sagiv2022personal}, and allows comparison to human samples. We then sorted these $v_i$ and ranked accordingly. Finally, we calculated the Spearman's Rank Correlation ($\rho$) between this ranking and the known human ranking \citep{schwartz2022measuring}. 
We note that this analysis does not consider correlations between answers given in the same session, and thus it may be viewed as analyzing the first-order statistics of the responses.

\subsubsection{Correlations Between Values \label{sec:correlations}}
A key focus of our work is correlation between values. Namely, the question of whether choice of value $i$ is correlated with that of value $j$. In humans there is a robust correlation structure where certain values are more strongly correlated than others. 
A standard way to represent this structure is via Multidimensional Scaling (MDS) \citep{borg2018applied}, calculated as follows. 

First, the matrix $C\in R^{19\times 19}$ of empirical correlation coefficients is formed. Next each of the values is embedded into $R^2$ via MDS, such that distances in $R^{2}$ best approximate the correlations. For human data, this results in an approximately circular embedding, as shown in \citep{schwartz2022measuring, skimina2021between, daniel2019value}. Here we performed this analysis on the LLM data. To compare the resulting dataset to the human samples, we need to normalize for the degrees of freedom of rotation and translation. This is done via Procrustes Analysis between the human and LLM embeddings. The resulting embeddings were plotted. Then, we computed the sum of squared differences between the procrusted MDS locations of each value to the human benchmark. Larger differences indicate stronger divergence from the human samples.

\section{Results} 
The above analyses were performed for all models and prompting strategies. We checked that model responses only contained scores for the questions in the questionnaires, and that they could therefore be transformed to tabular form and analyzed. This was almost always the case except for Gemma 2 27B on the Demographic prompt at temperature zero, and we therefore do not provide result for that settings.

\paragraph{Value Rankings:}
As previously mentioned, research across samples and cultures have shown that while individual differences exist in human value priorities, there are also robust common patterns. In this section, we analyze the LLM responses and compare them to the typical ranking of human values, as discussed in \secref{sec:rankings}. 

\begin{figure}[t]
    \centering
    \begin{minipage}{0.5\linewidth} 
        \centering
        \begin{subfigure}{\linewidth}
            \includegraphics[width=\linewidth]{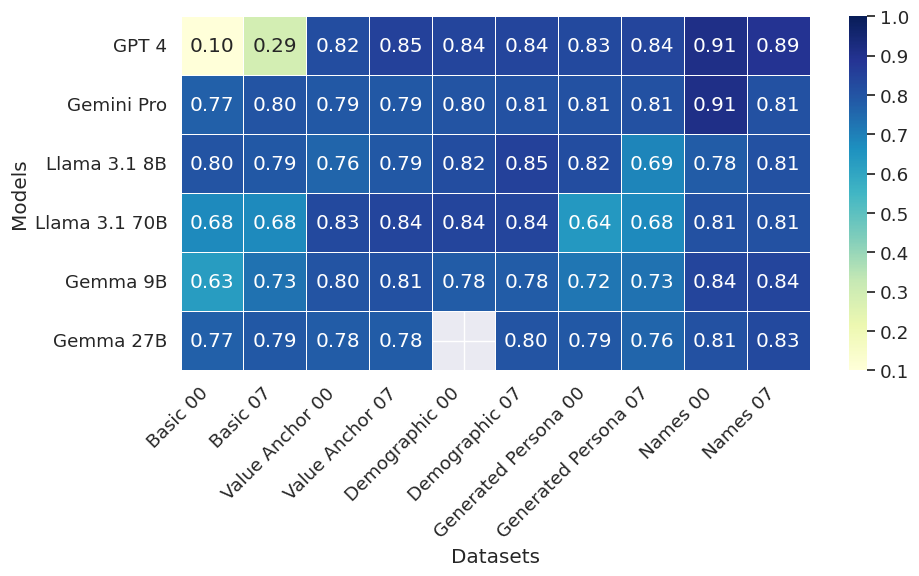}
            \caption{Spearman Rank Correlation Coefficients}
            \label{fig:CorHeatmap}
        \end{subfigure}
    \end{minipage}%
    \hspace{0.01\linewidth}
    \begin{minipage}{0.45\linewidth} 
        \centering
        \begin{subfigure}{\linewidth}
            \includegraphics[width=\linewidth]{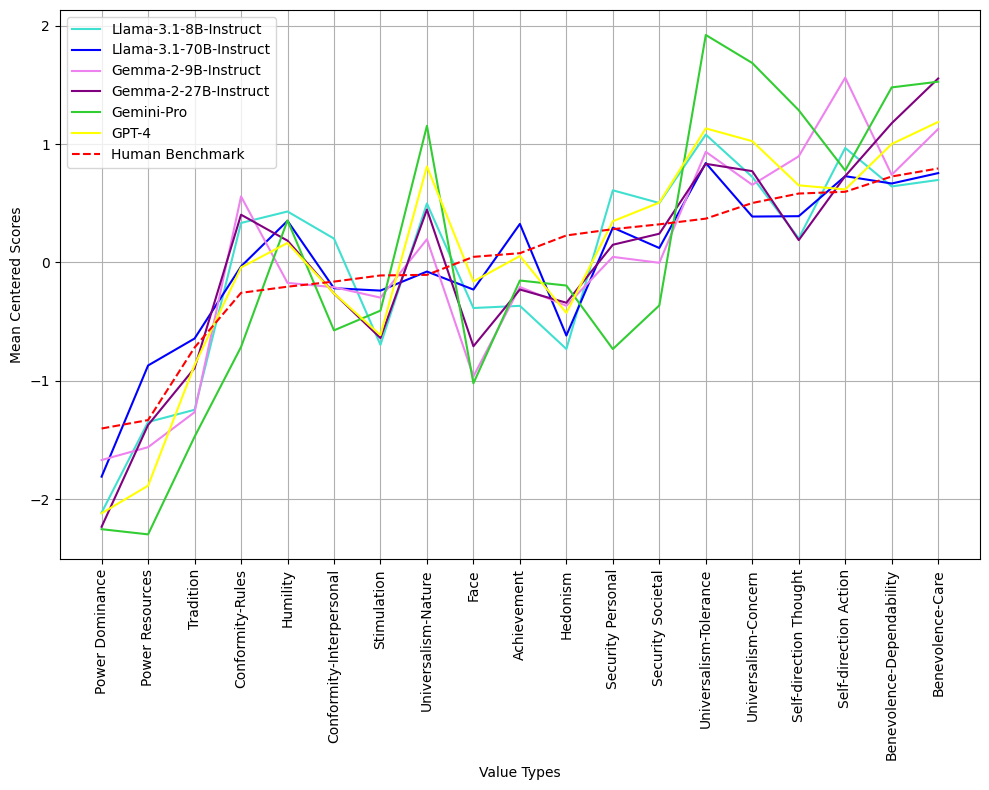}
            \caption{Value Ranking}
            \label{fig:ValueRankingSorts}
            \captionsetup[sub]{labelformat=empty} 
        \end{subfigure}
    \end{minipage}
    
    \caption{{\small {\bf Left:} A heatmap of Spearman rank correlation between benchmark value hierarchies and dataset rankings for GPT 4, Gemini Pro, Llama 3.1 8B and 70B instruct, and Gemma 2 9B and 27B across temperature conditions.
    {\bf Right:} Average value scores for the Value Anchoring prompt at zero temperature. The x-axis shows values ordered according to human ranking (i.e., Power ranks lowest for humans and Benevolence ranks highest). The y-axis is the mean-centered scores the models ascribe to these values in the questionnaire, and human values in red. It can be seen that models tend to give lower scores to values that are ranked lower by humans, and higher scores to values ranked higher. The LLM scores also track the human scores (red curve) quite well.  
    }}
    \captionsetup[sub]{labelformat=empty} 
    \label{fig:ranking_correlations}
\end{figure}

\figref{fig:CorHeatmap} shows the Spearman rank correlations between human rankings and those of the different models and prompting schemes. The results show high correlation levels ($>0.8$) for many prompt-model combinations. One exception is the basic prompt with GPT, which shows very low correlation. Full rankings are provided in the Appendix for several models and prompts (\tabref{tab:combined_GPT4_GeminiPro}, \tabref{tab:combined_Llama} and \tabref{tab:combined_Gemma}). These reveal that values such as Benevolence that are highly ranked in humans are indeed also highly ranked by most LLMs (e.g,. ranked third and first by GPT-4 for the Value Anchor prompt with temperatures $0$ and $0.7$ respectively). Conversely, values such as Power Dominance that are ranked low by humans, are ranked low by models (e.g., 19 by GPT-4 for the Value Anchor prompt). \figref{fig:ValueRankingSorts} shows the scores corresponding to the Value Anchor prompt, when sorted according to human preferences. It can be seen that the models tend to agree with the human ordering on the low and high ranked values. Taken together, these results demonstrate that LLMs tend on average to align with the human ranking of values. 


\comment{
The full rankings for Value and Name anchors are provided in the Appendix (see \tabref{tab:combined_GPT4_GeminiPro}, \tabref{tab:combined_Llama} and \tabref{tab:combined_Gemma}, which present the value rankings of humans and LLMs across two of the prompting methods - the Value Anchor and Names).

These tables reveal that LLM rankings generally align with human rankings in terms of which values are considered high or low. Specifically, values such as universalism, self-direction, and benevolence received high rankings, while values like tradition and power were consistently deemed less important.
}

\begin{figure}[t]
    \centering
    \begin{minipage}{0.42\linewidth} 
        \centering
        \begin{subfigure}{\linewidth}
            \includegraphics[width=\linewidth]{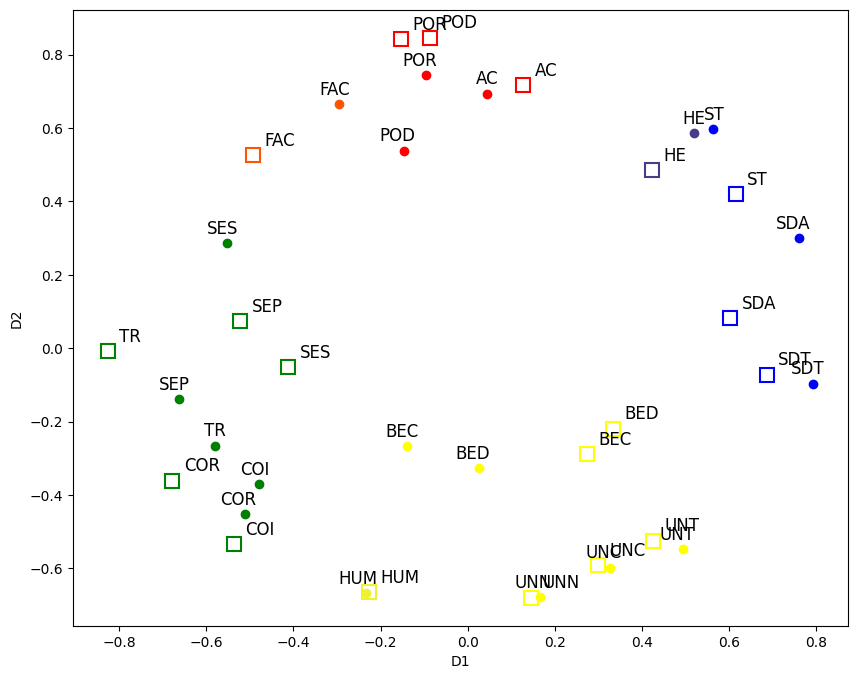}
            \caption{Value Anchor}
            \label{fig:Procrustes_BWVr_00}
        \end{subfigure}
    \end{minipage}%
    \begin{minipage}{0.16\linewidth} 
        \centering
        \includegraphics[width=\linewidth]{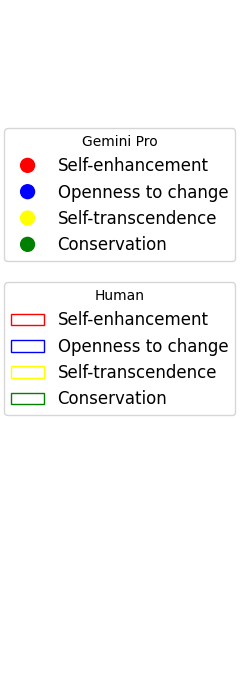}
    \end{minipage}%
    \begin{minipage}{0.42\linewidth} 
        \centering
        \begin{subfigure}{\linewidth}
            \includegraphics[width=\linewidth]{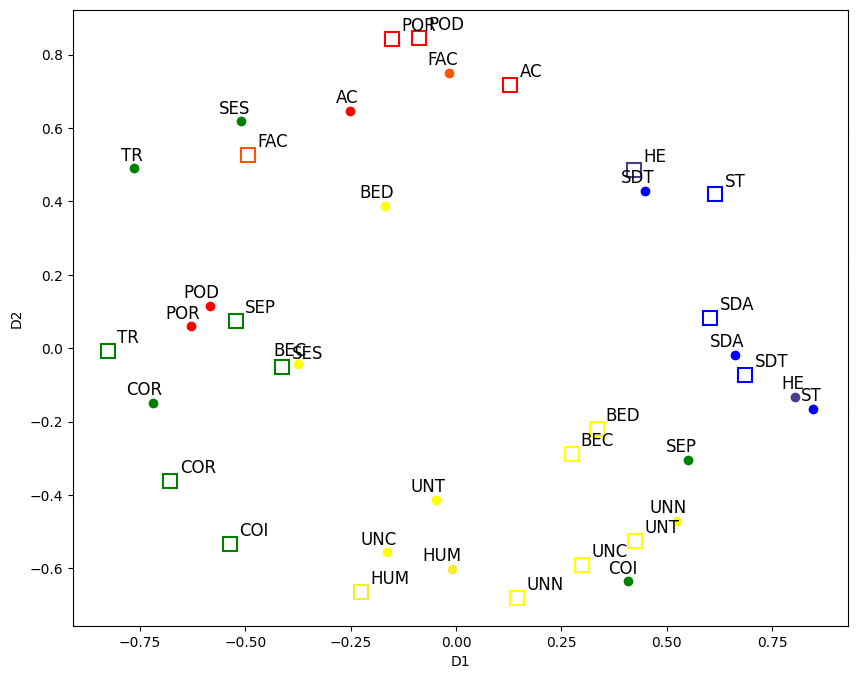}
            \caption{Names}
            \label{fig:Procrustes_names_00}
            \captionsetup[sub]{labelformat=empty} 
        \end{subfigure}
    \end{minipage}
    
    \caption{{\small Comparison of Procrustes Analysis results between human data \citep{schwartz2022measuring} and Gemini Pro for Value Anchor and Names prompts, for temperature $0.0$. The sum of squared differences, which measures the fit to human data, is 0.11 for the Value Anchor and 0.71 for the Names, indicating a better fit for the Value Anchor. For acronyms, refer to \secref{app:acronym}.}}
    \captionsetup[sub]{labelformat=empty} 
    \label{fig:Procrustes_names_BWVr_00}
\end{figure}

\paragraph{Correlations Between Values}
The MDS analysis (see \secref{sec:correlations}) maps all values into $R^2$ in a way that reflects their correlations. Here, we conduct MDS analyses for both human responses and LLM output, and then compare the results. The analyses were performed separately for each prompt, temperature, and model. \figref{fig:Procrustes_names_BWVr_00} illustrates a comparison between human MDS and Gemini Pro at temperature $0.0$, for the Value Anchor and Names prompts, respectively. Notably, the disparities between Gemini Pro and GPT-4 for each prompt are minimal. GPT-4 plots and other Gemini Pro prompts are included in the Appendix as \figref{fig:GeminiPro_demographic_persona} and \figref{fig:GPT4_all_prompts}.

First, it can be seen that among humans, the values are organized in a circle in the theoretically expected order. These results were often identified over the years, and interpreted as resulting from the aspiration of individuals to maintain personal consistency in their motivations \citep{schwartz1992universals}. Second, it can be seen that the MDS configuration resulting from the Value Anchor prompt more closely follows this order than the MDS resulting from the Names prompt.
We further quantitatively compared the configurations in \tabref{tab:sum square scores matrices}, by taking the mean squared difference between any pair of human and prompting method MDS matrices (i.e., matrices in $\mathbb{R}^{19\times 2}$). It can be seen that the Value Anchor prompt demonstrated a better fit to human values than the other prompting methods.

\begin{table}[t]
\centering
\begin{tabular}{@{}llllll@{}}
\toprule
 & \multicolumn{1}{c}{Basic} & \multicolumn{1}{c}{Value Anchor} & \multicolumn{1}{c}{Demographic} & \multicolumn{1}{c}{ Persona} & \multicolumn{1}{c}{Names}  \\
\midrule
GPT 4 & & & & & \\
00 & {0.92} & {\bf 0.23} & 0.53 & 0.25 & 0.32 \\
07 & {0.88} & {\bf 0.22} & 0.74 & {\bf 0.22} & 0.28 \\
\midrule
Gemini Pro & & & & & \\
00 & {0.87} & {\bf 0.11} & 0.42 & 0.39 & 0.71 \\
07 &   0.69         & {\bf 0.11} & 0.75 & 0.28 & 0.57 \\
\midrule
Llama 3.1 8B  & & & & & \\
00 & {0.80} & {\bf 0.18} & 0.47 & 0.58 & 0.60 \\
07 &   0.57         & {\bf 0.16} & 0.47 & 0.58 & 0.57 \\
\midrule
Llama 3.1 70B & & & & & \\
00 & {0.61} & {\bf 0.10} & 0.29 & 0.37 & 0.45 \\
07 &   0.44       & {\bf 0.10} & 0.22 & 0.40 & 0.44 \\
\midrule
Gemma 2 9B & & & & & \\
00 & {0.42} & {\bf 0.10} & 0.19 & 0.39 & 0.23 \\
07 &    0.82        & {\bf 0.11} & 0.16 & 0.32 & 0.12 \\
\midrule
Gemma 2 27B & & & & & \\
00 &  NA     & {\bf 0.16} & NA & 0.31 & 0.23 \\
07 &   0.64     & 0.17 & {\bf 0.15} & 0.25 & 0.19 \\
\bottomrule
\end{tabular}
\caption{Sum of squared difference between the MDS embeddings of humans and LLM. Gemma 2 27B did not produce parseable results for the Demographic prompt, and for Gemma 27B at temperature 0, some values had zero-variance, thus precluding computation of correlation coefficients. All Llamma models are Instruct.}
\label{tab:sum square scores matrices}
\end{table}

\subsection{Understanding Value Anchoring}
The results above show that value anchoring generates a correlation structure between values that is in better agreement with that of humans. We next set out to understand why this is the case. 
As we shall show, anchoring on a value not only increases the score the model assigns to the value, but it only changes the way other values are scored. Specifically, values that are close to the anchored value (on the value circle, \figref{fig:Theorized circle}) tend to receive high scores. On the other hand, values that are far from the anchor receive consistently lower scores.
This in turn has the effect of increasing correlation between values in the anchoring setting.

\begin{figure}[t]
    \centering
    \includegraphics[width=0.6\linewidth]{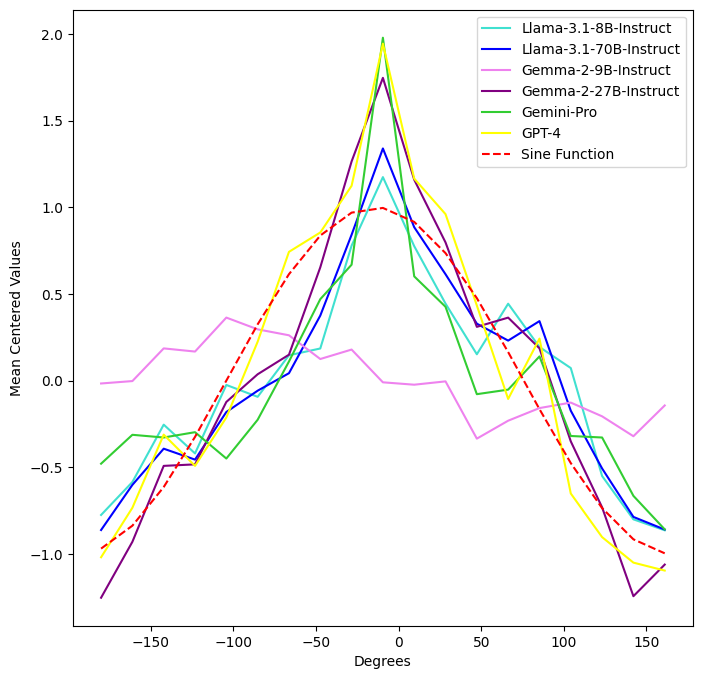}
    \caption{Analysis of scores after value anchoring. The plot shows the average of the score values after shifting to the anchored value. It can be seen that the anchored value receives the highest score, as expected. More surprisingly, neighboring values receive similarly high values, whereas more distant values receive lower values.}
    \label{fig:ideal_curve}
\end{figure}

To show the above, we produce an ``anchored score curve'' as follows. First, we order the 19 anchoring values according to their order on the value circle in \figref{fig:Theorized circle}. Note that in this order, values 19 and 1 will actually be close in the circle. Then, for each set of responses of an LLM to a Value Anchor prompt, we shift the anchored value to zero. We then average all these shifted curves. Results for all models are shown in \figref{fig:ideal_curve}, along with a sine function which shows a good fit to these curves. The behavior of all models is quite consistent, except for Gemma-2-9B.

As expected, the anchored value receives the highest score. However, what is more interesting is that the values close to it tend to get similarly high, and values farther away (e.g., 180 degrees apart) receive the lower values. This means that the anchored model scores values in a way that is consistent with its anchoring. This in turn implies that neighboring values will tend to be correlated, thus explaining why Value Anchoring better captures human correlation patterns. 

\section{Discussion}
In the current study we analyzed the values exhibited in LLM responses. We used two metrics to estimate the quality of the LLM responses against human responses: value ranking, and value correlations. 
Our results highlighted the importance of the prompting mechanism. Using the Basic prompt (namely, just providing a questionnaire with no further instructions), the LLM was likely to either generate negligible variance across generated personas, or generate internally inconsistent outputs (respond differently to questions about the same value). These results suggest that LLMs cannot be treated as 'individuals' holding a coherent set of value priorities. In contrast, our results indicated high consistency across model types, including commercial and open LLMs. 

Prompts that endow the LLM with a ``personality'' improved the consistency of each specific value profile, to varying degrees. The value hierarchy was consistently found across prompts, indicating that at the mean level, LLMs can simulate value rankings of human populations.
More variability between conditions was found in the 
measure of inter-value correlations. This is arguably the most important metric since it allows analyzing consistency across values, within a single session. We found best consistency in the Values Anchor prompt. These results suggest that LLMs, applying suitable prompts, can produce a 'population' of individuals, each reporting a different, but coherent set of value priorities. 

It is important to note that in neither of the prompts did the LLMs receive instructions for answering about all values. The results suggest that LLM is not only instruction-following, but uses the instructions as a context that guides consistent answers with relations to a variety of values. 
One fascinating question is where the LLM learns to produce such clear profiles of values. These profiles may be implicitly learned during pre-training. Indeed, past studies indicated that values can be identified in texts, such as newspaper articles and social media. However, these values did not necessary follow the theoretical value inter-relations identified here \citep{bardi2008new, ponizovskiy2020development, kumar2018inducing}. Individuals who value competing values may experience stress and indecision when faced with a dilemma, resulting in gradual change in values toward a more coherent form \citep{bardi2009structure, daniel2019value}. In contrast, text may very well present both sides of a dilemma and thus retain inconsistencies. Such inconsistencies were not identified because past studies relied mostly on lexical approaches \citep{bardi2008new, ponizovskiy2020development, kumar2018inducing}. LLMs, taking context into account, may be more likely to identify value inter-relations correctly. LLMs may also have learned to produce value profiles in the process of fine-tuning or RLHF \citep{qiu2022towards}. Future research can try to distinguish these two sources of learning using careful analysis of training sources as well as evaluation of different checkpoints in the training process. 

\comment{
The literature debates how LLMs can be used to simulate human behavior. Some studies apply the approach of 'LLM as an individual' \citep[e.g.,][]{binz2023using, stevenson2022putting}. Here, this approach (i.e., Basic prompt 00)  succeeded in replicating value ranking but failed on other measures. Other studies consider the 'LLM as a population' approach, where different prompts result in variability across responses \citep{argyle2023out}. 
Our Basic 07, Value anchor, Demographic, Generated Persona, and Names prompts, all took this approach, with differing success. 
}
Past research into human personas sought ways to estimate the ability of the LLM to maintain a consistent persona across a conversation \citep{wang2024incharacter}. We establish that the unique qualities of human values, and the ample empirical knowledge collected about them, allow their use as such a method \citep{sagiv2022personal, sagiv2017personal, knafo2024development}. We suggest that known behavioral correlations in humans can be applied to assess the consistency of LLM personas. Here we focused on evaluations via a questionnaire, but one could envision more elaborate evaluations that rely on other features of human personalities. 

The procedures and data produced here may have important contributions for psychological research. Investigators interested in human behavior can apply these procedures to produce datasets that simulate human samples. Future research can investigate their possible use to replicate known findings (e.g., age differences in values) or pretest novel hypotheses (e.g. associations between values and specific behaviors). The use of both commercial and open LLMs increases the reproducibilty of the results. 

The question of values in LLMs is of course of philosophical and societal importance. Our results show that on  average, these values largely reproduce international value rankings. However, small variations in value importance may have implications at the societal and individual level (e.g. gender roles \citep{lomazzi2020gender}; entrepreneurship \citep{woodside2020consequences}; prosocial behavior \citep{daniel2020value}; and antisocial behavior \citep{benish2019can}). Future work should consider the influence of these values on LLM responses, as well as on the individuals interacting with them. 

The current study focused on a limited number of contexts (e.g., five prompts, two temperatures, and six models). Importantly, we found commonalities across the various contexts, beside the differences. Future studies can use these results to understand what other contexts should be investigated, to possibly further enhance the quality of the output. Another limitation is the restriction to one questionnaire (PVQ-RR) corresponding a specific value system. It will be interesting to explore other forms of probing values. 

\bibliographystyle{unsrtnat}

\begin{sidewaystable}[!ht]
\centering
\caption{Comparative analysis of 19 values' relative importance of the Value Anchor and Names datasets across temperatures for GPT-4 and Gemini Pro.}
\label{tab:combined_GPT4_GeminiPro}
\small{
\begin{tabular}{@{}lllllllllllllllllll@{}}
\toprule
\multicolumn{3}{c}{Benchmark} & \multicolumn{8}{c}{GPT-4} & \multicolumn{8}{c}{Gemini Pro} \\
\cmidrule(lr){4-11} \cmidrule(l){12-19}
\multicolumn{3}{c}{Human Data} & \multicolumn{2}{c}{Value Anchor 00} & \multicolumn{2}{c}{Value Anchor 07} & \multicolumn{2}{c}{Names 00} & \multicolumn{2}{c}{Names 07} & \multicolumn{2}{c}{Value Anchor 00} & \multicolumn{2}{c}{Value Anchor 07} & \multicolumn{2}{c}{Names 00} & \multicolumn{2}{c}{Names 07} \\
\cmidrule(r){1-3} \cmidrule(lr){4-5} \cmidrule(lr){6-7} \cmidrule(lr){8-9} \cmidrule(l){10-11} \cmidrule(lr){12-13} \cmidrule(lr){14-15} \cmidrule(lr){16-17} \cmidrule(l){18-19}
Rank  & Values & Mean & Mean & Rank & Mean & Rank & Mean & Rank & Mean & Rank & Mean & Rank & Mean & Rank & Mean & Rank & Mean & Rank \\
\midrule
1 & BEC & 0.79 & 1.18 & 3 & 1.24 & 1 & 0.66 & 3 & 0.65 & 3.5 & 1.52 & 3 & 1.44 & 3 & 1.56 & 3 & -0.10 & 13 \\
2 & BED & 0.72 & 0.92 & 5 & 1.00 & 4 & 0.66 & 3 & 0.65 & 6 & 1.39 & 4 & 1.30 & 4 & 1.48 & 5 & 0.03 & 12 \\
3 & SDA & 0.60 & 0.59 & 7 & 0.55 & 7 & 0.66 & 3 & 0.65 & 3.5 & 0.88 & 7 & 0.79 & 7 & 1.20 & 7 & 0.93 & 1 \\
4 & SDT & 0.58 & 0.70 & 6 & 0.69 & 6 & 0.66 & 3 & 0.65 & 1.5 & 1.33 & 5 & 1.17 & 5 & 1.48 & 4 & 0.47 & 5 \\
5 & UNC & 0.50 & 1.12 & 3 & 1.07 & 3 & 0.66 & 3 & 0.65 & 1.5 & 1.68 & 2 & 1.63 & 2 & 1.68 & 1 & 0.74 & 3 \\
6 & UNT & 0.37 & 1.26 & 1 & 1.20 & 2 & 0.66 & 6 & 0.65 & 5 & 1.93 & 1 & 1.86 & 1 & 1.63 & 2 & 0.45 & 6 \\
7 & SES & 0.32 & 0.41 & 8 & 0.36 & 8 & 0.64 & 7 & 0.58 & 7 & -0.49 & 12 & -0.42 & 12 & 0.60 & 8 & 0.14 & 10 \\
8 & SEP & 0.28 & 0.21 & 9 & 0.26 & 9 & 0.48 & 11 & 0.43 & 9 & -0.87 & 14 & -0.75 & 14 & -0.75 & 14 & 0.21 & 8 \\
9 & HE & 0.23 & -0.31 & 14 & -0.33 & 14 & 0.50 & 9 & 0.41 & 10 & 0.04 & 9 & -0.02 & 9 & 0.04 & 10 & 0.57 & 4 \\
10 & AC & 0.08 & 0.07 & 11 & 0.09 & 11 & 0.06 & 14 & 0.10 & 14 & -0.08 & 11 & -0.22 & 11 & -0.35 & 11 & -0.54 & 16 \\
11 & FAC & 0.05 & -0.28 & 13 & -0.31 & 13 & 0.30 & 12 & 0.20 & 13 & -1.01 & 16 & -0.91 & 16 & -0.68 & 13 & 0.10 & 11 \\
12 & UNN & -0.10 & 0.98 & 4 & 0.97 & 4 & 0.58 & 8 & 0.51 & 8 & 1.17 & 6 & 1.10 & 6 & 1.29 & 6 & 0.19 & 9 \\
13 & ST & -0.11 & -0.48 & 16 & -0.44 & 16 & -0.10 & 15 & -0.14 & 15 & -0.07 & 10 & -0.09 & 10 & -0.68 & 12 & -1.03 & 18 \\
14 & COI & -0.16 & -0.41 & 15 & -0.42 & 15 & -0.74 & 17 & -0.65 & 17 & -0.72 & 13 & -0.58 & 13 & -0.90 & 16 & 0.34 & 7 \\
15 & HUM & -0.20 & 0.20 & 10 & 0.20 & 10 & 0.30 & 13 & 0.22 & 12 & 0.12 & 8 & 0.15 & 8 & 0.52 & 9 & 0.84 & 2 \\
16 & COR & -0.26 & -0.21 & 12 & -0.18 & 12 & 0.48 & 10 & 0.41 & 11 & -0.95 & 15 & -0.79 & 15 & -1.18 & 17 & -0.33 & 14 \\
17 & TR & -0.72 & -0.98 & 17 & -0.93 & 17 & -0.62 & 16 & -0.59 & 16 & -1.57 & 17 & -1.43 & 17 & -0.88 & 15 & -0.52 & 15 \\
18 & POR & -1.33 & -1.91 & 18 & -1.83 & 18 & -2.60 & 18 & -2.58 & 18 & -2.19 & 19 & -2.12 & 19 & -3.09 & 19 & -0.83 & 17 \\
19 & POD & -1.40 & -2.14 & 19 & -2.05 & 19 & -2.82 & 19 & -2.81 & 19 & -2.14 & 18 & -2.10 & 18 & -2.98 & 18 & -1.64 & 19 \\
\bottomrule
\end{tabular}
}
\end{sidewaystable}

\begin{sidewaystable}[!ht]
\centering
\caption{Comparative analysis of values' relative importance for the Llama 3.1 8B and Llama 3.1 70B datasets across temperatures.}
\label{tab:combined_Llama}
\small{
\begin{tabular}{@{}lllllllllllllllllll@{}}
\toprule
\multicolumn{3}{c}{Benchmark} & \multicolumn{8}{c}{Llama 3.1 8B} & \multicolumn{8}{c}{Llama 3.1 70B} \\
\cmidrule(lr){4-11} \cmidrule(l){12-19}
\multicolumn{3}{c}{Human Data} & \multicolumn{2}{c}{Value Anchor 00} & \multicolumn{2}{c}{Value Anchor 07} & \multicolumn{2}{c}{Names 00} & \multicolumn{2}{c}{Names 07} & \multicolumn{2}{c}{Value Anchor 00} & \multicolumn{2}{c}{Value Anchor 07} & \multicolumn{2}{c}{Names 00} & \multicolumn{2}{c}{Names 07} \\
\cmidrule(r){1-3} \cmidrule(lr){4-5} \cmidrule(lr){6-7} \cmidrule(lr){8-9} \cmidrule(l){10-11} \cmidrule(lr){12-13} \cmidrule(lr){14-15} \cmidrule(lr){16-17} \cmidrule(l){18-19}
Rank  & Values & Mean & Mean & Rank & Mean & Rank & Mean & Rank & Mean & Rank & Mean & Rank & Mean & Rank & Mean & Rank & Mean & Rank \\
\midrule
1 & BEC & 0.79 & 0.74 & 4 & 0.87 & 3 & 0.62 & 8 & 0.82 & 4 & 0.80 & 2 & 0.80 & 2 & 0.87 & 3 & 0.83 & 5 \\
2 & BED & 0.72 & 0.64 & 5 & 0.64 & 5 & 0.62 & 9 & 0.65 & 7 & 0.64 & 4 & 0.69 & 4 & 0.67 & 7 & 0.69 & 7 \\
3 & SDA & 0.60 & 1.00 & 2 & 1.04 & 2 & 1.12 & 2 & 1.12 & 2 & 0.76 & 3 & 0.72 & 3 & 0.84 & 5 & 0.84 & 4 \\
4 & SDT & 0.58 & 0.24 & 11 & 0.34 & 9 & 0.89 & 5 & 0.67 & 6 & 0.48 & 5 & 0.53 & 5 & 1.01 & 2 & 0.96 & 2 \\
5 & UNC & 0.50 & 0.77 & 3 & 0.80 & 4 & 1.11 & 3 & 0.99 & 3 & 0.46 & 6 & 0.49 & 6 & 0.85 & 4 & 0.84 & 3 \\
6 & UNT & 0.37 & 1.14 & 1 & 1.12 & 1 & 1.19 & 1 & 1.20 & 1 & 0.90 & 1 & 0.88 & 1 & 1.10 & 1 & 1.10 & 1 \\
7 & SES & 0.32 & 0.34 & 9 & 0.47 & 8 & 0.91 & 4 & 0.81 & 5 & -0.01 & 11 & 0.01 & 10 & 0.05 & 11 & 0.08 & 11 \\
8 & SEP & 0.28 & 0.62 & 6 & 0.58 & 6 & 0.84 & 6 & 0.64 & 8 & 0.23 & 9 & 0.23 & 9 & 0.34 & 8 & 0.39 & 8 \\
9 & HE & 0.23 & -0.61 & 16 & -0.58 & 16 & -1.38 & 16 & -0.50 & 14 & -0.51 & 16 & -0.46 & 16 & -0.13 & 13 & -0.31 & 13 \\
10 & AC & 0.08 & -0.37 & 13 & -0.36 & 13 & -0.27 & 13 & -1.20 & 12 & 0.32 & 8 & 0.26 & 8 & 0.23 & 10 & 0.24 & 9 \\
11 & FAC & 0.05 & -0.46 & 14 & -0.58 & 15 & 0.75 & 15 & -0.63 & 16 & -0.24 & 14 & -0.35 & 14 & -0.60 & 15 & -0.66 & 15 \\
12 & UNN & -0.10 & 0.55 & 7 & 0.53 & 7 & 0.56 & 11 & 0.48 & 9 & 0.01 & 10 & -0.03 & 11 & 0.29 & 9 & 0.15 & 10 \\
13 & ST & -0.11 & -0.59 & 15 & -0.53 & 14 & -0.67 & 14 & -0.34 & 13 & -0.15 & 13 & -0.12 & 13 & -0.03 & 12 & -0.08 & 12 \\
14 & COI & -0.16 & -0.00 & 12 & -0.14 & 12 & 0.61 & 10 & 0.29 & 10 & -0.30 & 15 & -0.37 & 15 & -0.89 & 17 & -0.74 & 16 \\
15 & HUM & -0.20 & 0.44 & 8 & 0.29 & 10 & 0.69 & 7 & 0.47 & 10 & 0.35 & 7 & 0.29 & 7 & 0.74 & 6 & 0.81 & 6 \\
16 & COR & -0.26 & 0.29 & 10 & 0.22 & 11 & 0.56 & 12 & 0.03 & 11 & -0.14 & 12 & -0.11 & 12 & -0.35 & 14 & -0.36 & 14 \\
17 & TR & -0.72 & -1.28 & 17 & -1.28 & 17 & -2.01 & 18 & -1.53 & 17 & -0.68 & 17 & -0.66 & 17 & -0.82 & 16 & -0.82 & 17 \\
18 & POR & -1.33 & -1.35 & 18 & -1.29 & 18 & -1.77 & 17 & -1.55 & 18 & -0.89 & 18 & -0.88 & 18 & -1.46 & 18 & -1.46 & 18 \\
19 & POD & -1.40 & -2.12 & 19 & -2.11 & 19 & -2.88 & 19 & -2.56 & 19 & -1.86 & 19 & -1.81 & 19 & -2.39 & 19 & -2.36 & 19 \\
\bottomrule
\end{tabular}
}
\end{sidewaystable}

\begin{sidewaystable}[!ht]
\centering
\caption{Comparative analysis of values' relative importance for the Gemma 2 9B and Gemma 2 27B datasets across temperatures.}
\label{tab:combined_Gemma}
\small{
\begin{tabular}{@{}lllllllllllllllllll@{}}
\toprule
\multicolumn{3}{c}{Benchmark} & \multicolumn{8}{c}{Gemma 2 9B} & \multicolumn{8}{c}{Gemma 2 27B} \\
\cmidrule(lr){4-11} \cmidrule(l){12-19}
\multicolumn{3}{c}{Human Data} & \multicolumn{2}{c}{Value Anchor 00} & \multicolumn{2}{c}{Value Anchor 07} & \multicolumn{2}{c}{Names 00} & \multicolumn{2}{c}{Names 07} & \multicolumn{2}{c}{Value Anchor 00} & \multicolumn{2}{c}{Value Anchor 07} & \multicolumn{2}{c}{Names 00} & \multicolumn{2}{c}{Names 07} \\
\cmidrule(r){1-3} \cmidrule(lr){4-5} \cmidrule(lr){6-7} \cmidrule(lr){8-9} \cmidrule(l){10-11} \cmidrule(lr){12-13} \cmidrule(lr){14-15} \cmidrule(lr){16-17} \cmidrule(l){18-19}
Rank  & Values & Mean & Mean & Rank & Mean & Rank & Mean & Rank & Mean & Rank & Mean & Rank & Mean & Rank & Mean & Rank & Mean & Rank \\
\midrule
1 & BEC & 0.79 & 1.14 & 2 & 1.22 & 2 & 1.90 & 2 & 1.88 & 2 & 1.57 & 1 & 1.57 & 1 & 1.53 & 1 & 1.50 & 1 \\
2 & BED & 0.72 & 0.73 & 5 & 0.78 & 5 & 1.29 & 4 & 1.24 & 4 & 1.16 & 2 & 1.17 & 2 & 1.29 & 3 & 1.25 & 3 \\
3 & SDA & 0.60 & 1.55 & 1 & 1.58 & 1 & 1.77 & 3 & 1.77 & 3 & 0.84 & 4 & 0.84 & 4 & 1.50 & 2 & 1.46 & 2 \\
4 & SDT & 0.58 & 0.90 & 4 & 0.89 & 3 & 1.93 & 1 & 1.91 & 1 & 0.30 & 8 & 0.27 & 8 & 1.20 & 4 & 1.18 & 4 \\
5 & UNC & 0.50 & 0.67 & 6 & 0.65 & 6 & 1.03 & 6 & 1.05 & 5 & 0.83 & 5 & 0.81 & 5 & 1.02 & 6 & 1.04 & 5 \\
6 & UNT & 0.37 & 0.94 & 3 & 0.84 & 4 & 1.03 & 5 & 1.02 & 6 & 0.92 & 3 & 0.89 & 3 & 1.02 & 5 & 1.02 & 6 \\
7 & SES & 0.32 & -0.01 & 10 & 0.00 & 10 & 0.75 & 8 & 0.76 & 8 & 0.11 & 10 & 0.12 & 10 & 0.32 & 10 & 0.33 & 9 \\
8 & SEP & 0.28 & 0.08 & 9 & 0.05 & 9 & 0.15 & 9 & 0.12 & 9 & 0.07 & 11 & 0.07 & 11 & 0.11 & 11 & 0.07 & 11 \\
9 & HE & 0.23 & -0.33 & 15 & -0.32 & 15 & -0.54 & 13 & -0.55 & 13 & -0.23 & 13 & -0.23 & 13 & -0.55 & 13 & -0.49 & 13 \\
10 & AC & 0.08 & -0.22 & 13 & -0.22 & 13 & -0.50 & 12 & -0.50 & 12 & -0.19 & 12 & -0.21 & 12 & -0.34 & 12 & -0.32 & 12 \\
11 & FAC & 0.05 & -1.01 & 16 & -1.05 & 16 & -1.45 & 17 & -1.45 & 17 & -0.84 & 16 & -0.81 & 16 & -1.24 & 17 & -1.24 & 17 \\
12 & UNN & -0.10 & 0.21 & 8 & 0.20 & 8 & -0.03 & 10 & 0.01 & 10 & 0.56 & 6 & 0.45 & 6 & 0.55 & 7 & 0.61 & 7 \\
13 & ST & -0.11 & -0.27 & 14 & -0.24 & 14 & -0.72 & 14 & -0.77 & 14 & -0.49 & 15 & -0.42 & 15 & -0.72 & 14 & -0.77 & 15 \\
14 & COI & -0.16 & -0.21 & 12 & -0.20 & 12 & -0.94 & 15 & -0.93 & 15 & -0.33 & 14 & -0.31 & 14 & -1.00 & 16 & -1.01 & 16 \\
15 & HUM & -0.20 & -0.14 & 11 & -0.18 & 11 & -0.40 & 11 & -0.41 & 11 & 0.13 & 9 & 0.13 & 9 & 0.47 & 8 & 0.47 & 8 \\
16 & COR & -0.26 & 0.55 & 7 & 0.63 & 7 & 0.84 & 7 & 0.85 & 7 & 0.30 & 7 & 0.29 & 7 & 0.34 & 9 & 0.31 & 10 \\
17 & TR & -0.72 & -1.23 & 17 & -1.24 & 17 & -0.99 & 16 & -0.96 & 16 & -0.99 & 17 & -0.96 & 17 & -0.80 & 15 & -0.76 & 14 \\
18 & POR & -1.33 & -1.63 & 18 & -1.62 & 18 & -2.53 & 18 & -2.48 & 18 & -1.41 & 18 & -1.41 & 18 & -2.00 & 18 & -1.96 & 18 \\
19 & POD & -1.40 & -1.72 & 19 & -1.77 & 19 & -2.59 & 19 & -2.57 & 19 & -2.30 & 19 & -2.26 & 19 & -2.70 & 19 & -2.69 & 19 \\
\bottomrule
\end{tabular}
}
\end{sidewaystable}

\bibliography{references}

\begin{thebibliography}{52}
\providecommand{\natexlab}[1]{#1}
\providecommand{\url}[1]{\texttt{#1}}
\expandafter\ifx\csname urlstyle\endcsname\relax
  \providecommand{\doi}[1]{doi: #1}\else
  \providecommand{\doi}{doi: \begingroup \urlstyle{rm}\Url}\fi

\bibitem[Aher et~al.(2023)Aher, Arriaga, and Kalai]{aher2023using}
Gati~V Aher, Rosa~I Arriaga, and Adam~Tauman Kalai.
\newblock Using large language models to simulate multiple humans and replicate human subject studies.
\newblock In \emph{International Conference on Machine Learning}, pages 337--371. Proceedings of Machine Learning Research, 2023.

\bibitem[Fischer et~al.(2023)Fischer, Luczak-Roesch, and Karl]{fischer2023what}
Ronald Fischer, Markus Luczak-Roesch, and Johannes~A Karl.
\newblock What does chatgpt return about human values? exploring value bias in chatgpt using a descriptive value theory.
\newblock \emph{arXiv preprint}, 2023.

\bibitem[Sagiv and Schwartz(2022)]{sagiv2022personal}
Lilach Sagiv and Shalom~H Schwartz.
\newblock Personal values across cultures.
\newblock \emph{Annual review of psychology}, 73\penalty0 (1):\penalty0 517--546, 2022.
\newblock \doi{10.1146/annurev-psych-020821-125100}.

\bibitem[Sagiv et~al.(2017)Sagiv, Roccas, Cieciuch, and Schwartz]{sagiv2017personal}
Lilach Sagiv, Sonia Roccas, Jan Cieciuch, and Shalom~H Schwartz.
\newblock Personal values in human life.
\newblock \emph{Nature human behaviour}, 1\penalty0 (9):\penalty0 630--639, 2017.
\newblock \doi{10.1038/s41562-017-0185-3}.

\bibitem[Roberts and Yoon(2022)]{roberts2022personality}
Brent~W Roberts and Hee~J Yoon.
\newblock Personality psychology.
\newblock \emph{Annual Review of Psychology}, 73\penalty0 (1):\penalty0 489--516, 2022.
\newblock \doi{10.1146/annurev-psych-020821-114927}.

\bibitem[Schwartz(1992)]{schwartz1992universals}
Shalom~H Schwartz.
\newblock Universals in the content and structure of values: Theoretical advances and empirical tests in 20 countries.
\newblock In \emph{Advances in experimental social psychology}, volume~25, pages 1--65. Elsevier, 1992.

\bibitem[Schwartz(2012)]{schwartz2012overview}
Shalom~H Schwartz.
\newblock An overview of the {S}chwartz theory of basic values.
\newblock \emph{Online readings in Psychology and Culture}, 2\penalty0 (1):\penalty0 1--20, 2012.
\newblock \doi{10.9707/2307-0919.1116}.

\bibitem[Schwartz and Bardi(2001)]{schwartz2001value}
Shalom~H Schwartz and Anat Bardi.
\newblock Value hierarchies across cultures: Taking a similarities perspective.
\newblock \emph{Journal of cross-cultural Psychology}, 32\penalty0 (3):\penalty0 268--290, 2001.
\newblock \doi{10.1177/0022022101032003002}.

\bibitem[Skimina et~al.(2021{\natexlab{a}})Skimina, Cieciuch, and Revelle]{skimina2021between}
Ewa Skimina, Jan Cieciuch, and William Revelle.
\newblock Between-and within-person structures of value traits and value states: Four different structures, four different interpretations.
\newblock \emph{Journal of Personality}, 89\penalty0 (5):\penalty0 951--969, 2021{\natexlab{a}}.

\bibitem[Daniel et~al.(2023)Daniel, D{\"o}ring, and Cieciuch]{daniel2023development}
Ella Daniel, Anna~K D{\"o}ring, and Jan Cieciuch.
\newblock Development of intraindividual value structures in middle childhood: A multicultural and longitudinal investigation.
\newblock \emph{Journal of Personality}, 91\penalty0 (2):\penalty0 482--496, 2023.

\bibitem[Schwartz and Cieciuch(2022)]{schwartz2022measuring}
Shalom~H Schwartz and Jan Cieciuch.
\newblock Measuring the refined theory of individual values in 49 cultural groups: psychometrics of the revised portrait value questionnaire.
\newblock \emph{Assessment}, 29\penalty0 (5):\penalty0 1005--1019, 2022.
\newblock \doi{10.1177/1073191121998760}.

\bibitem[Schwartz et~al.(2012)Schwartz, Cieciuch, Vecchione, Davidov, Fischer, Beierlein, Ramos, Verkasalo, L{\"o}nnqvist, Demirutku, et~al.]{schwartz2012refining}
Shalom~H Schwartz, Jan Cieciuch, Michele Vecchione, Eldad Davidov, Ronald Fischer, Constanze Beierlein, Alice Ramos, Markku Verkasalo, Jan-Erik L{\"o}nnqvist, Kursad Demirutku, et~al.
\newblock Refining the theory of basic individual values.
\newblock \emph{Journal of personality and social psychology}, 103\penalty0 (4):\penalty0 663--688, 2012.
\newblock \doi{10.1037/a0029393}.

\bibitem[Schwartz(2017)]{schwartz2017refined}
Shalom~H Schwartz.
\newblock The refined theory of basic values.
\newblock \emph{Values and behavior: Taking a cross cultural perspective}, pages 51--72, 2017.

\bibitem[Pakizeh et~al.(2007)Pakizeh, Gebauer, and Maio]{pakizeh2007basic}
Ali Pakizeh, Jochen~E Gebauer, and Gregory~R Maio.
\newblock Basic human values: Inter-value structure in memory.
\newblock \emph{Journal of Experimental Social Psychology}, 43\penalty0 (3):\penalty0 458--465, 2007.
\newblock \doi{10.1016/j.jesp.2006.04.007}.

\bibitem[Skimina et~al.(2021{\natexlab{b}})Skimina, Cieciuch, and Strus]{skimina2021traits}
Ewa Skimina, Jan Cieciuch, and W{\l}odzimierz Strus.
\newblock Traits and values as predictors of the frequency of everyday behavior: Comparison between models and levels.
\newblock \emph{Current Psychology}, 40\penalty0 (1):\penalty0 133--153, 2021{\natexlab{b}}.
\newblock \doi{10.1007/s12144-018-9892-9}.

\bibitem[Lindahl and Saeid(2023)]{lindahl2023unveiling}
Caroline Lindahl and Helin Saeid.
\newblock Unveiling the values of {ChatGPT}: An explorative study on human values in {AI} systems, 2023.

\bibitem[Miotto et~al.(2022)Miotto, Rossberg, and Kleinberg]{miotto2022gpt}
Maril{\`u} Miotto, Nicola Rossberg, and Bennett Kleinberg.
\newblock Who is {GPT}-3? an exploration of personality, values and demographics.
\newblock \emph{arXiv}, 2022.

\bibitem[Scherrer et~al.(2023)Scherrer, Shi, Feder, and Blei]{ScherrerSFB23}
Nino Scherrer, Claudia Shi, Amir Feder, and David~M. Blei.
\newblock Evaluating the moral beliefs encoded in llms.
\newblock In \emph{Advances in Neural Information Processing Systems 36: Annual Conference on Neural Information Processing Systems 2023, NeurIPS 2023, New Orleans, LA, USA, December 10 - 16, 2023}, 2023.

\bibitem[Hadar-Shoval et~al.(2024)Hadar-Shoval, Asraf, Mizrachi, Haber, and Elyoseph]{hadar2024assessing}
Dorit Hadar-Shoval, Kfir Asraf, Yonathan Mizrachi, Yuval Haber, and Zohar Elyoseph.
\newblock Assessing the alignment of large language models with human values for mental health integration: Cross-sectional study using schwartz’s theory of basic values.
\newblock \emph{JMIR Mental Health}, 11, 2024.

\bibitem[Kova{\v{c}} et~al.(2023)Kova{\v{c}}, Sawayama, Portelas, Colas, Dominey, and Oudeyer]{kovavc2023large}
Grgur Kova{\v{c}}, Masataka Sawayama, R{\'e}my Portelas, C{\'e}dric Colas, Peter~Ford Dominey, and Pierre-Yves Oudeyer.
\newblock Large language models as superpositions of cultural perspectives.
\newblock \emph{arXiv preprint arXiv:2307.07870}, 2023.

\bibitem[Liu et~al.(2023)Liu, Yuan, Fu, Jiang, Hayashi, and Neubig]{liu2023pre}
Pengfei Liu, Weizhe Yuan, Jinlan Fu, Zhengbao Jiang, Hiroaki Hayashi, and Graham Neubig.
\newblock Pre-train, prompt, and predict: A systematic survey of prompting methods in natural language processing.
\newblock \emph{ACM Computing Surveys}, 55\penalty0 (9):\penalty0 1--35, 2023.
\newblock \doi{10.1145/3560815}.

\bibitem[Hadar-Shoval et~al.(2023)Hadar-Shoval, Elyoseph, and Lvovsky]{hadar2023plasticity}
Dorit Hadar-Shoval, Zohar Elyoseph, and Maya Lvovsky.
\newblock The plasticity of chatgpt’s mentalizing abilities: Personalization for personality structures.
\newblock \emph{Frontiers in Psychiatry}, 14:\penalty0 1234397, 2023.
\newblock \doi{10.3389/fpsyt.2023.1234397}.

\bibitem[Jiang et~al.(2023)Jiang, Zhang, Cao, Kabbara, and Roy]{jiang2023personallm}
Hang Jiang, Xiajie Zhang, Xubo Cao, Jad Kabbara, and Deb Roy.
\newblock Personallm: Investigating the ability of gpt-3.5 to express personality traits and gender differences.
\newblock \emph{arXiv}, 2023.

\bibitem[Salewski et~al.(2024)Salewski, Alaniz, Rio-Torto, Schulz, and Akata]{salewski2024context}
Leonard Salewski, Stephan Alaniz, Isabel Rio-Torto, Eric Schulz, and Zeynep Akata.
\newblock In-context impersonation reveals large language models' strengths and biases.
\newblock \emph{Advances in Neural Information Processing Systems}, 36, 2024.

\bibitem[Argyle et~al.(2023)Argyle, Busby, Fulda, Gubler, Rytting, and Wingate]{argyle2023out}
Lisa~P Argyle, Ethan~C Busby, Nancy Fulda, Joshua~R Gubler, Christopher Rytting, and David Wingate.
\newblock Out of one, many: Using language models to simulate human samples.
\newblock \emph{Political Analysis}, 31\penalty0 (3):\penalty0 337--351, 2023.

\bibitem[Safdari et~al.(2023)Safdari, Serapio-Garc{\'\i}a, Crepy, Fitz, Romero, Sun, Abdulhai, Faust, and Matari{\'c}]{safdari2023personality}
Mustafa Safdari, Greg Serapio-Garc{\'\i}a, Cl{\'e}ment Crepy, Stephen Fitz, Peter Romero, Luning Sun, Marwa Abdulhai, Aleksandra Faust, and Maja Matari{\'c}.
\newblock Personality traits in large language models.
\newblock \emph{arXiv preprint arXiv:2307.00184}, 2023.

\bibitem[Li et~al.(2023)Li, Chong, Stepputtis, Campbell, Hughes, Lewis, and Sycara]{li2023theory}
Huao Li, Yu~Quan Chong, Simon Stepputtis, Joseph Campbell, Dana Hughes, Michael Lewis, and Katia Sycara.
\newblock Theory of mind for multi-agent collaboration via large language models.
\newblock \emph{arXiv preprint arXiv:2310.10701}, 2023.

\bibitem[Gunel et~al.(2020)Gunel, Du, Conneau, and Stoyanov]{gunel2020supervised}
Beliz Gunel, Jingfei Du, Alexis Conneau, and Ves Stoyanov.
\newblock Supervised contrastive learning for pre-trained language model fine-tuning.
\newblock \emph{arXiv preprint arXiv:2011.01403}, 2020.

\bibitem[Hagendorff et~al.(2023)Hagendorff, Fabi, and Kosinski]{hagendorff2023human}
Thilo Hagendorff, Sarah Fabi, and Michal Kosinski.
\newblock Human-like intuitive behavior and reasoning biases emerged in large language models but disappeared in chatgpt.
\newblock \emph{Nature Computational Science}, 3\penalty0 (10):\penalty0 833--838, 2023.
\newblock \doi{10.1038/s43588-023-00527-x}.

\bibitem[Binz and Schulz(2023)]{binz2023using}
Marcel Binz and Eric Schulz.
\newblock Using cognitive psychology to understand gpt-3.
\newblock \emph{Proceedings of the National Academy of Sciences}, 120\penalty0 (6), 2023.
\newblock \doi{10.1073/pnas.2218523120}.

\bibitem[Ouyang et~al.(2022)Ouyang, Wu, Jiang, Almeida, Wainwright, Mishkin, Zhang, Agarwal, Slama, Ray, et~al.]{ouyang2022training}
Long Ouyang, Jeffrey Wu, Xu~Jiang, Diogo Almeida, Carroll Wainwright, Pamela Mishkin, Chong Zhang, Sandhini Agarwal, Katarina Slama, Alex Ray, et~al.
\newblock Training language models to follow instructions with human feedback.
\newblock \emph{Advances in neural information processing systems}, 35:\penalty0 27730--27744, 2022.

\bibitem[Stevenson et~al.(2022)Stevenson, Smal, Baas, Grasman, and van~der Maas]{stevenson2022putting}
Claire Stevenson, Iris Smal, Matthijs Baas, Raoul Grasman, and Han van~der Maas.
\newblock Putting gpt-3's creativity to the (alternative uses) test.
\newblock \emph{arXiv preprint arXiv:2206.08932}, 2022.

\bibitem[Deshpande et~al.(2023)Deshpande, Murahari, Rajpurohit, Kalyan, and Narasimhan]{deshpande2023toxicity}
Ameet Deshpande, Vishvak Murahari, Tanmay Rajpurohit, Ashwin Kalyan, and Karthik Narasimhan.
\newblock Toxicity in chatgpt: Analyzing persona-assigned language models.
\newblock \emph{arXiv preprint}, 2023.

\bibitem[Wang et~al.(2024)Wang, Xiao, tse Huang, Yuan, Xu, Guo, Tu, Fei, Leng, Wang, Chen, Li, and Xiao]{wang2024incharacter}
Xintao Wang, Yunze Xiao, Jen tse Huang, Siyu Yuan, Rui Xu, Haoran Guo, Quan Tu, Yaying Fei, Ziang Leng, Wei Wang, Jiangjie Chen, Cheng Li, and Yanghua Xiao.
\newblock Incharacter: Evaluating personality fidelity in role-playing agents through psychological interviews.
\newblock \emph{arXiv preprint arXiv:2310.17976}, 2024.

\bibitem[Gupta et~al.(2024)Gupta, Shrivastava, Deshpande, Kalyan, Clark, Sabharwal, and Khot]{gupta2024bias}
Shashank Gupta, Vaishnavi Shrivastava, Ameet Deshpande, Ashwin Kalyan, Peter Clark, Ashish Sabharwal, and Tushar Khot.
\newblock Bias runs deep: Implicit reasoning biases in persona-assigned llms.
\newblock \emph{arXiv preprint arXiv:2311.04892}, 2024.

\bibitem[Lee et~al.(2019)Lee, Sneddon, Daly, Schwartz, Soutar, and Louviere]{lee2019testing}
Julie~A Lee, Joanne~N Sneddon, Timothy~M Daly, Shalom~H Schwartz, Geoffrey~N Soutar, and Jordan~J Louviere.
\newblock Testing and extending schwartz refined value theory using a best--worst scaling approach.
\newblock \emph{Assessment}, 26\penalty0 (2):\penalty0 166--180, 2019.
\newblock \doi{10.1177/1073191116683799}.

\bibitem[{National Academies of Sciences, Engineering, and Medicine}(2022)]{national2022measuring}
{National Academies of Sciences, Engineering, and Medicine}.
\newblock \emph{Measuring Sex, Gender Identity, and Sexual Orientation}.
\newblock National Academies Press, Washington, DC, 2022.

\bibitem[Haerpfer et~al.(2022)Haerpfer, Inglehart, Moreno, Welzel, Kizilova, Diez-Medrano, Lagos, Norris, Ponarin, and Puranen]{haerpfer2022world}
Christian Haerpfer, Ronald Inglehart, Alejandro Moreno, Christian Welzel, Kseniya Kizilova, Jaime Diez-Medrano, Marta Lagos, Pippa Norris, Eduard Ponarin, and Bi~Puranen, editors.
\newblock \emph{World Values Survey: Round Seven - Country-Pooled Datafile Version 5.0}.
\newblock JD Systems Institute \& WVSA Secretariat, Madrid, Spain \& Vienna, Austria, 2022.
\newblock \doi{10.14281/18241.20}.

\bibitem[Laver-Fawcett et~al.(2016)Laver-Fawcett, Brain, Brodie, Cardy, and Manaton]{laver2016face}
Alison Laver-Fawcett, Leanne Brain, Courtney Brodie, Lauren Cardy, and Lisa Manaton.
\newblock The face validity and clinical utility of the activity card sort--united kingdom (acs-uk).
\newblock \emph{British Journal of Occupational Therapy}, 79\penalty0 (8):\penalty0 492--504, 2016.
\newblock \doi{10.1177/0308022616629167}.

\bibitem[Cheng et~al.(2023)Cheng, Durmus, and Jurafsky]{cheng2023marked}
Myra Cheng, Esin Durmus, and Dan Jurafsky.
\newblock Marked personas: Using natural language prompts to measure stereotypes in language models.
\newblock \emph{arXiv preprint arXiv:2305.18189}, 2023.

\bibitem[Borg et~al.(2018)Borg, Groenen, and Mair]{borg2018applied}
Ingwer Borg, Patrick~JF Groenen, and Patrick Mair.
\newblock \emph{Applied multidimensional scaling and unfolding}.
\newblock Springer Science \& Business Media, New York, NY, 2nd edition, 2018.
\newblock \doi{https://doi.org/10.1007/978-3-319-73471-2}.

\bibitem[Daniel and Benish-Weisman(2019)]{daniel2019value}
Ella Daniel and Maya Benish-Weisman.
\newblock Value development during adolescence: Dimensions of change and stability.
\newblock \emph{Journal of personality}, 87\penalty0 (3):\penalty0 620--632, 2019.

\bibitem[Bardi et~al.(2008)Bardi, Calogero, and Mullen]{bardi2008new}
Anat Bardi, Rachel~M Calogero, and Brian Mullen.
\newblock A new archival approach to the study of values and value--behavior relations: validation of the value lexicon.
\newblock \emph{Journal of Applied Psychology}, 93\penalty0 (3):\penalty0 483, 2008.

\bibitem[Ponizovskiy et~al.(2020)Ponizovskiy, Ardag, Grigoryan, Boyd, Dobewall, and Holtz]{ponizovskiy2020development}
Vladimir Ponizovskiy, Murat Ardag, Lusine Grigoryan, Ryan Boyd, Henrik Dobewall, and Peter Holtz.
\newblock Development and validation of the personal values dictionary: A theory--driven tool for investigating references to basic human values in text.
\newblock \emph{European Journal of Personality}, 34\penalty0 (5):\penalty0 885--902, 2020.

\bibitem[Kumar et~al.(2018)Kumar, Reganti, Maheshwari, Chakroborty, Gamb{\"a}ck, and Das]{kumar2018inducing}
Upendra Kumar, Aishwarya~N Reganti, Tushar Maheshwari, Tanmoy Chakroborty, Bj{\"o}rn Gamb{\"a}ck, and Amitava Das.
\newblock Inducing personalities and values from language use in social network communities.
\newblock \emph{Information Systems Frontiers}, 20:\penalty0 1219--1240, 2018.

\bibitem[Bardi et~al.(2009)Bardi, Lee, Hofmann-Towfigh, and Soutar]{bardi2009structure}
Anat Bardi, Julie~Anne Lee, Nadi Hofmann-Towfigh, and Geoffrey Soutar.
\newblock The structure of intraindividual value change.
\newblock \emph{Journal of personality and social psychology}, 97\penalty0 (5):\penalty0 913, 2009.

\bibitem[Qiu et~al.(2022)Qiu, Zhao, Liang, Lu, Shi, Yu, and Zhu]{qiu2022towards}
Liang Qiu, Yizhou Zhao, Yuan Liang, Pan Lu, Weiyan Shi, Zhou Yu, and Song-Chun Zhu.
\newblock Towards socially intelligent agents with mental state transition and human value.
\newblock In \emph{Proceedings of the 23rd Annual Meeting of the Special Interest Group on Discourse and Dialogue}, pages 146--158, 2022.

\bibitem[Knafo-Noam et~al.(2024)Knafo-Noam, Daniel, and Benish-Weisman]{knafo2024development}
Ariel Knafo-Noam, Ella Daniel, and Maya Benish-Weisman.
\newblock The development of values in middle childhood: Five maturation criteria.
\newblock \emph{Current Directions in Psychological Science}, 33\penalty0 (1):\penalty0 18--26, 2024.

\bibitem[Lomazzi and Seddig(2020)]{lomazzi2020gender}
Vera Lomazzi and Daniel Seddig.
\newblock Gender role attitudes in the international social survey programme: Cross-national comparability and relationships to cultural values.
\newblock \emph{Cross-Cultural Research}, 54\penalty0 (4):\penalty0 398--431, 2020.

\bibitem[Woodside et~al.(2020)Woodside, Megehee, Isaksson, and Ferguson]{woodside2020consequences}
Arch~G Woodside, Carol~M Megehee, Lars Isaksson, and Graham Ferguson.
\newblock Consequences of national cultures and motivations on entrepreneurship, innovation, ethical behavior, and quality-of-life.
\newblock \emph{Journal of Business \& Industrial Marketing}, 35\penalty0 (1):\penalty0 40--60, 2020.

\bibitem[Daniel et~al.(2020)Daniel, Benish-Weisman, Sneddon, and Lee]{daniel2020value}
Ella Daniel, Maya Benish-Weisman, Joanne~N Sneddon, and Julie~A Lee.
\newblock Value profiles during middle childhood: Developmental processes and social behavior.
\newblock \emph{Child Development}, 91\penalty0 (5):\penalty0 1615--1630, 2020.

\bibitem[Benish-Weisman(2019)]{benish2019can}
Maya Benish-Weisman.
\newblock What can we learn about aggression from what adolescents consider important in life? the contribution of values theory to aggression research.
\newblock \emph{Child Development Perspectives}, 13\penalty0 (4):\penalty0 260--266, 2019.

\end{thebibliography}
\clearpage

\appendix
\section{Additional Files}
The Python and R code used to generate our prompt sets and analyses can be made available upon request, and will be added to open-source
repositories for wider public use soon.

\section{Question Variants \label{sec:variants}}
For each value, we use three different question variants.
For example, the three question variants formulated to assess individuals' alignment with the value type "Power Dominance" (i.e., relating to asserting authority and control over others) are as follows:
\begin{itemize}
    \item Question 6: "He desires recognition for his abilities and seeks admiration for his actions."
    \item Question 29: "He prefers taking charge of situations and making decisions."
    \item Question 41: "He actively seeks positions of power and influence, valuing control and authority over others."
\end{itemize}

\section{Value Acronyms\label{app:acronym}}
The figures in the paper use the following value acronyms: SDT = Self-Direction Thought; SDA = Self-Direction Action; ST = Stimulation; HE = Hedonism; AC = Achievement; POD = Power-Dominance; POR = Power-Resources; FAC = Face; SEP = Security-Personal; SES = Security-Societal; TR = Tradition; COR = Conformity-Rules; COI = Conformity-Interpersonal; HUM = Humility; UNN = Universalism-Nature; UNC = Universalism-Concern; UNT = Universalism-Tolerance; BEC = Benevolence-Caring; BED = Benevolence-Dependability

\section{Example Portrait Value Questionnaire}
\begin{figure}[h]
    \centering
    \includegraphics[width=0.6\linewidth]{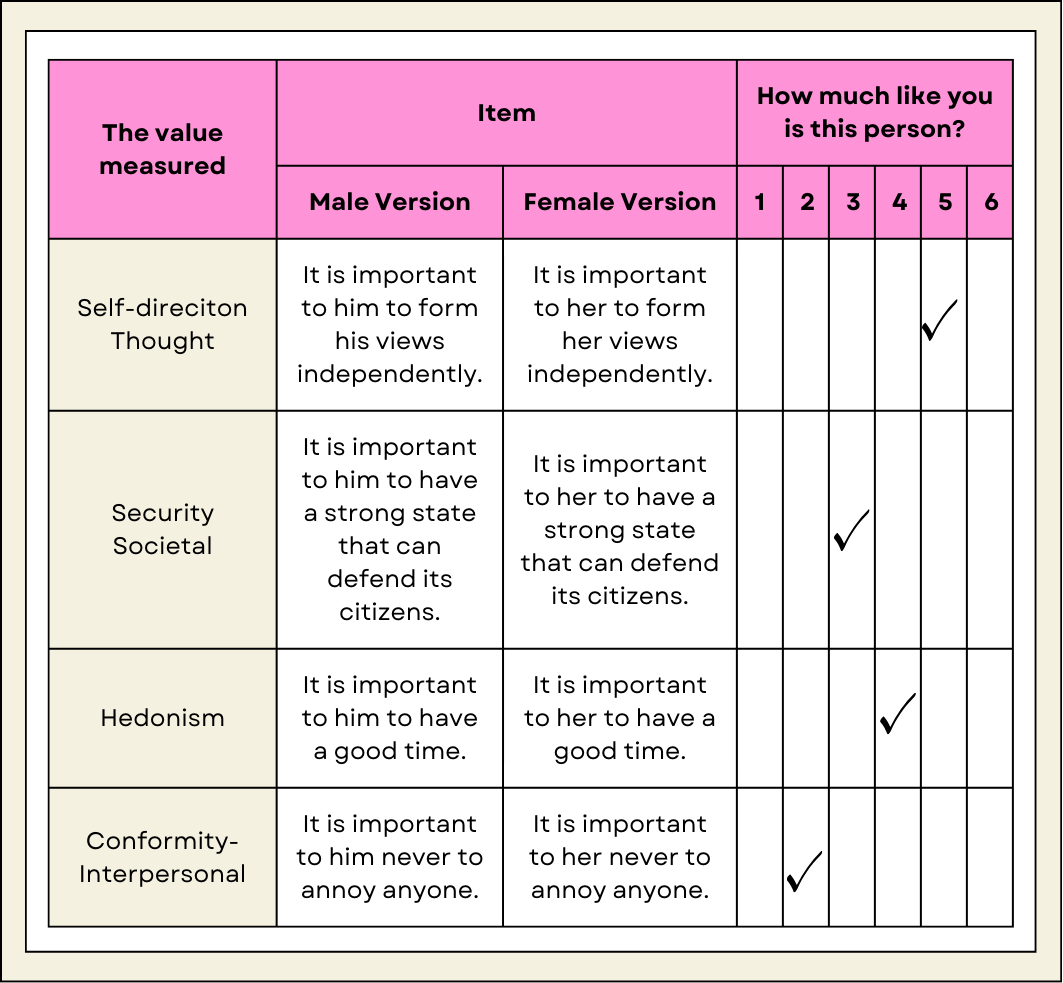}
    \caption{Portrait Value Questionnaire—Revised - example items. The instructions provided were: ``Here we briefly describe some people. Please read each description and think about how much each person is or is not like you. Tick the box to the right that shows how much the person in the description is like you''. Rankings correspond to the following descriptions: 1-Not like me at all, 2-Not like me, 3-A little like me, 4-Somewhat like me, 5-Like me, 6-Very much like me.}
    \label{fig:PVQ items}
\end{figure}

\figref{fig:PVQ items} provides an example for the Portrait Value Questionnaire that was used in our study. 

\section{The Complete Item List of Best-Worst Refined Values (BWVr)}
\label{sec:BWVr list}
In our value anchoring approach, we used the description of values in \cite{lee2019testing} to prompt the LLMs. The set of descriptions is provided below.
\begin{enumerate}[label={\textbf{\arabic*.}},align=left,leftmargin=*]

\item \textbf{Self-direction-thought}: developing your own original ideas and opinions
\item \textbf{Self-direction-action}: being free to act independently
\item \textbf{Stimulation}: having an exciting life; having all sorts of new experiences
\item \textbf{Hedonism}: taking advantage of every opportunity to enjoy life’s pleasures
\item \textbf{Achievement}: being ambitious and successful
\item \textbf{Power-dominance}: having the power that money and possessions can bring
\item \textbf{Power-resources}: having the authority to get others to do what you want
\item \textbf{Face}: protecting your public image and avoiding being shamed
\item \textbf{Security-personal}: living and acting in ways that ensure that you are personally safe and secure
\item \textbf{Security-societal}: living in a safe and stable society
\item \textbf{Tradition}: following cultural family or religious practices
\item \textbf{Conformity-rules}: obeying all rules and laws
\item \textbf{Conformity-interpersonal}: making sure you never upset or annoy others
\item \textbf{Humility}: being humble and avoiding public recognition
\item \textbf{Benevolence-dependability}: being a completely dependable and trustworthy friend and family member
\item \textbf{Benevolence-caring}: helping and caring for the wellbeing of those who are close
\item \textbf{Universalism-concern}: caring and seeking justice for everyone especially the weak and vulnerable in society
\item \textbf{Universalism-nature}: protecting the natural environment from destruction or pollution
\item \textbf{Universalism-tolerance}: being open-minded and accepting of people and ideas, even when you disagree with them
\item \textbf{Animal welfare}: caring for the welfare of animals
\end{enumerate}

\section{Additional Results for Value Rankings}
\tabref{tab:combined_GPT4_GeminiPro} provides additional results on value rankings for several prompting approaches. 


\comment{
\begin{figure}[!htb]
    \centering
    \includegraphics[width=\linewidth]{cronbach's_alpha_07_gpt4_gemini_llama_8_70_gemma_9_27.png}
    \caption{Comparing mean Cronbach's \(\alpha\) scores and standard deviations of LLMs in temperature $0.7$ across different categories with human data \citep{schwartz2022measuring}.}
    \label{fig:cronbach's alpha 07}
\end{figure}
}
\section{\texorpdfstring{Additional MDS Plots}{MDS figures for Gemini Pro and GPT 4 for temperature $0.0$}}
\begin{figure}[ht]
    \centering
    \begin{minipage}{0.5\linewidth}
        \centering
        \begin{subfigure}{\linewidth}
            \includegraphics[width=\linewidth]{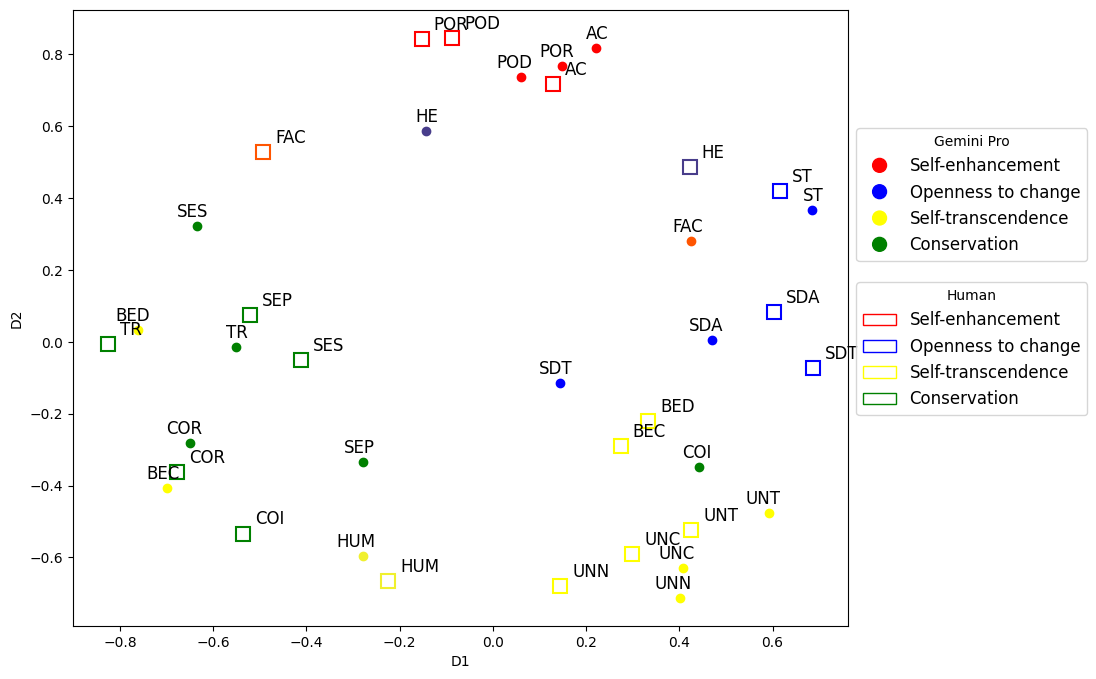}
            \caption{Demographic}
            \label{fig:Gemini_demographic00}
        \end{subfigure}
    \end{minipage}%
    \begin{minipage}{0.5\linewidth}
        \centering
        \begin{subfigure}{\linewidth}
            \includegraphics[width=\linewidth]{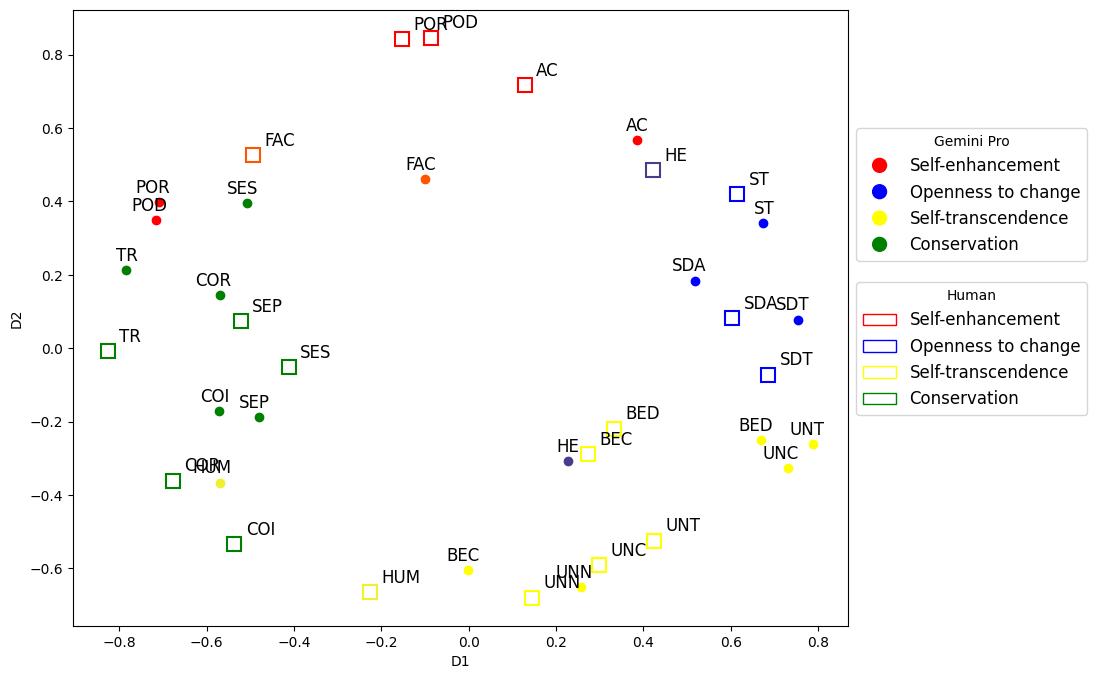}
            \caption{Generated Persona}
            \label{fig:Gemini_persona00}
            \captionsetup[sub]{labelformat=empty} 
        \end{subfigure}
    \end{minipage}
    \caption{Comparison of the MDS results between human data \citep{schwartz2022measuring} and Gemini Pro for Demographic and Generated Persona respectively, in the temperature $0.0$ condition.}
    \label{fig:GeminiPro_demographic_persona}
\end{figure}

\begin{figure}[ht]
    \centering
    \begin{minipage}{0.5\linewidth}
        \centering
        \begin{subfigure}{\linewidth}
            \includegraphics[width=\linewidth]{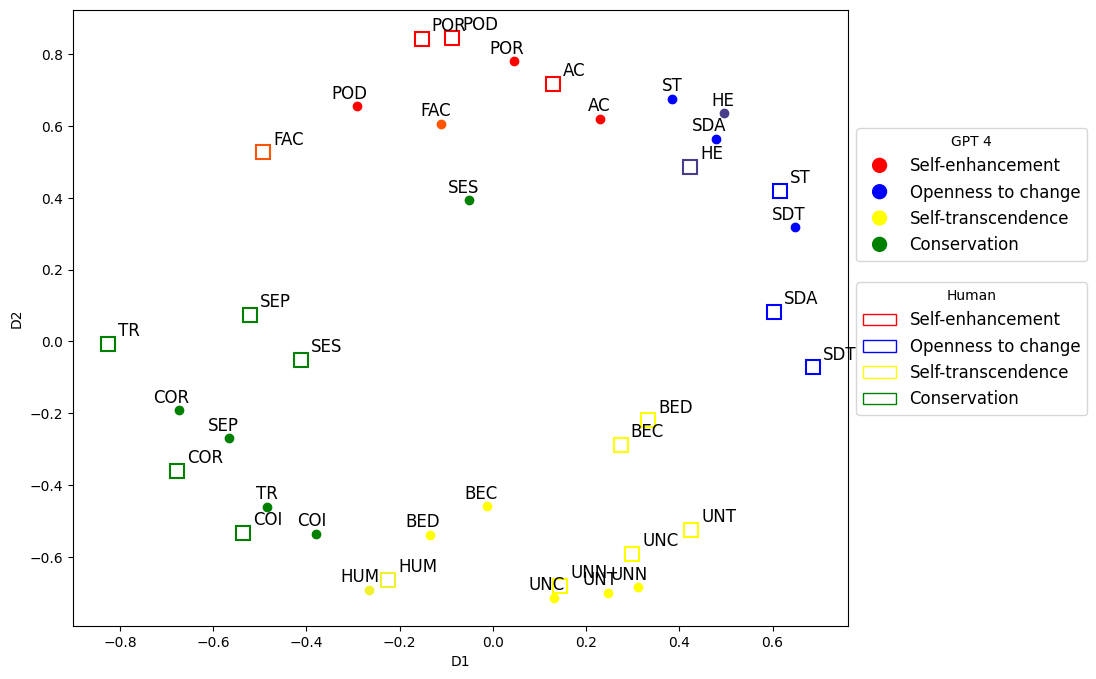}
            \caption{Value Anchor}
            \label{fig:GPT4_bwvr00}
        \end{subfigure}
        \begin{subfigure}{\linewidth}
            \includegraphics[width=\linewidth]{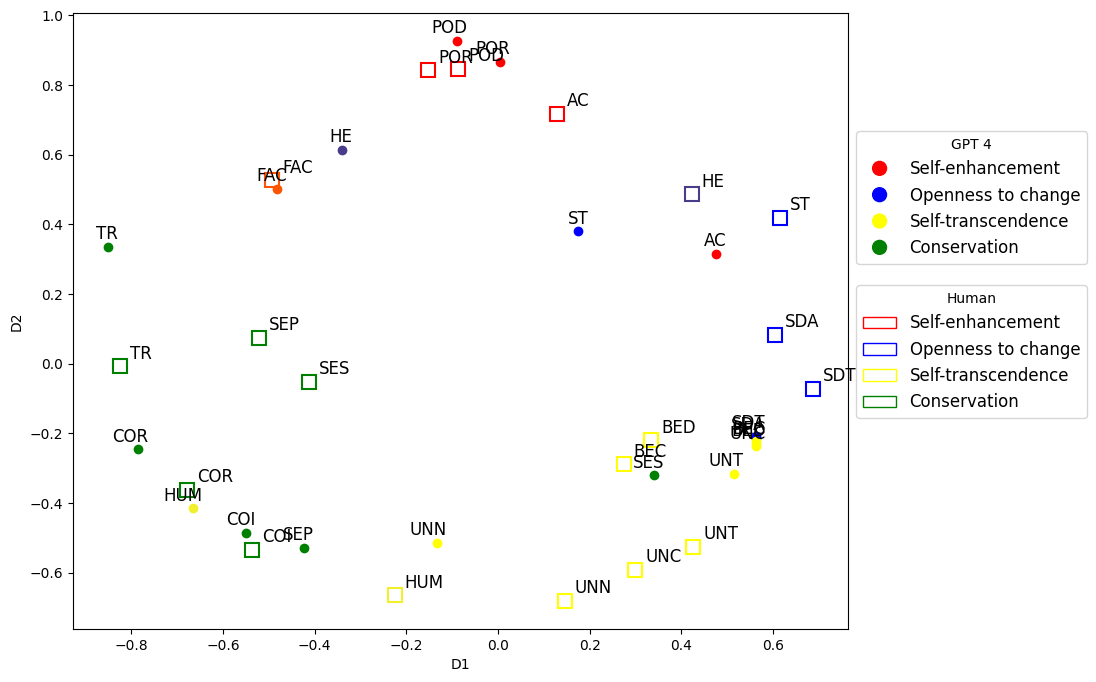}
            \caption{Names}
            \label{fig:GPT4_names00}
            \captionsetup[sub]{labelformat=empty} 
        \end{subfigure}
    \end{minipage}%
    \begin{minipage}{0.5\linewidth}
        \centering
        \begin{subfigure}{\linewidth}
            \includegraphics[width=\linewidth]{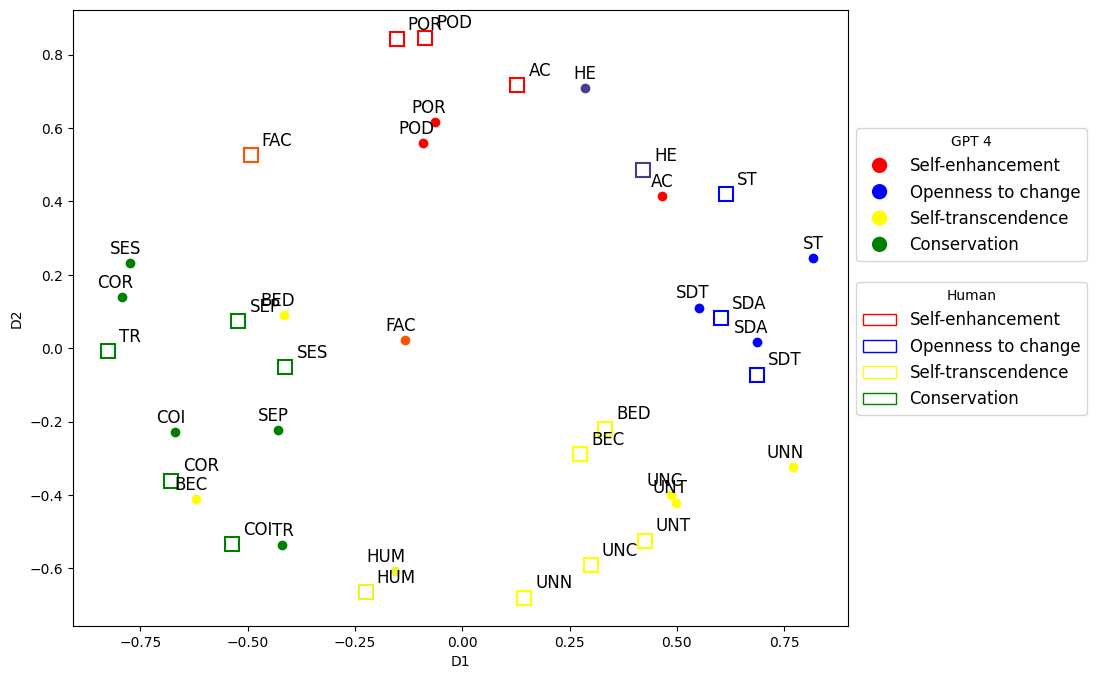}
            \caption{Demographic}
            \label{fig:GPT4_demographic00}
        \end{subfigure}
        \begin{subfigure}{\linewidth}
            \includegraphics[width=\linewidth]{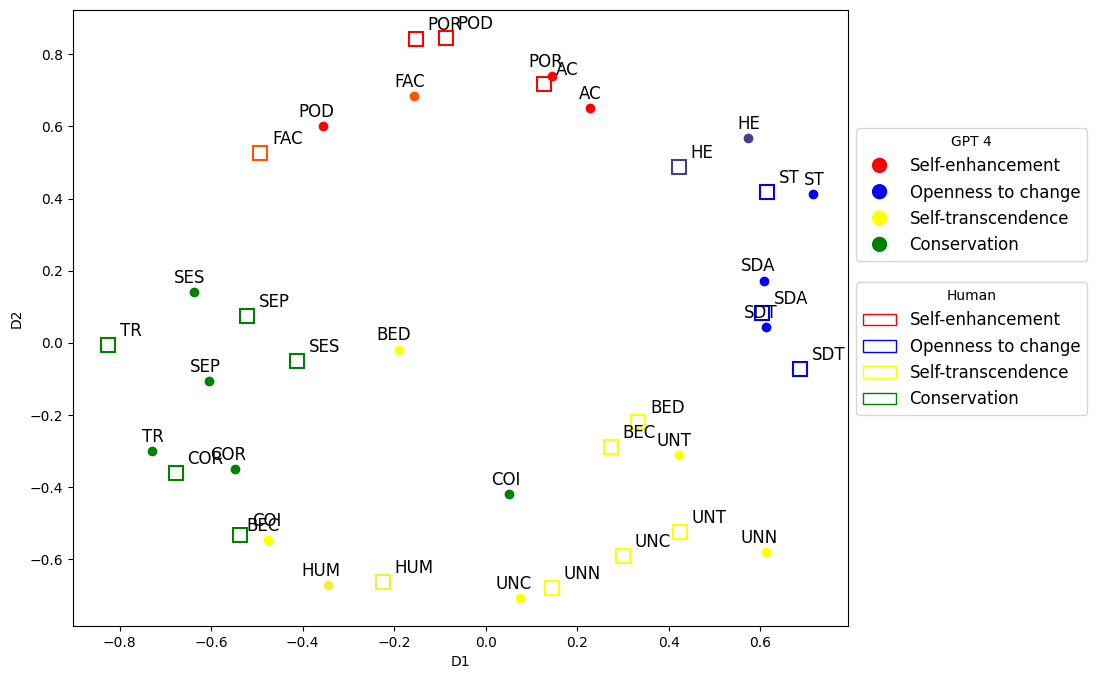}
            \caption{Generated Persona}
            \label{fig:GPT4_persona00}
        \end{subfigure}
    \end{minipage}
    \caption{Comparison of the MDS results between human data \citep{schwartz2022measuring} and GPT 4 for all prompts, in the temperature $0.0$ condition.}
    \label{fig:GPT4_all_prompts}
\end{figure}

\begin{figure}[ht]
    \centering
    \begin{minipage}{0.5\linewidth}
        \centering
        \begin{subfigure}{\linewidth}
            \includegraphics[width=\linewidth]{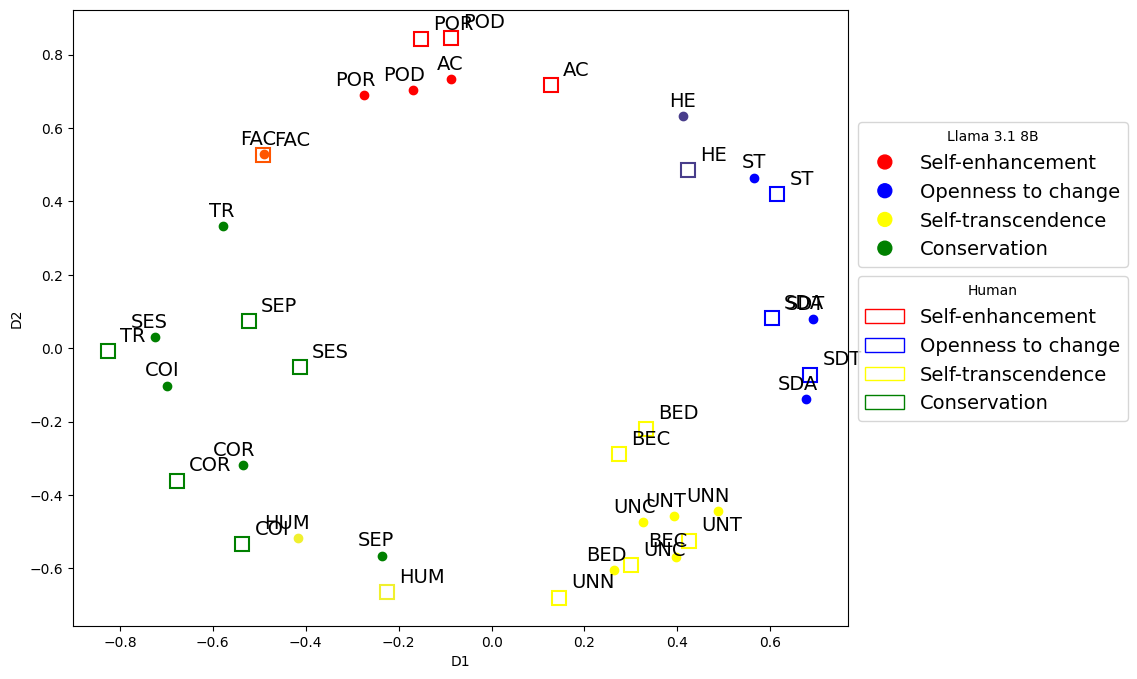}
            \caption{Value Anchor}
            \label{fig:Llama_bwvr00}
        \end{subfigure}
        \begin{subfigure}{\linewidth}
            \includegraphics[width=\linewidth]{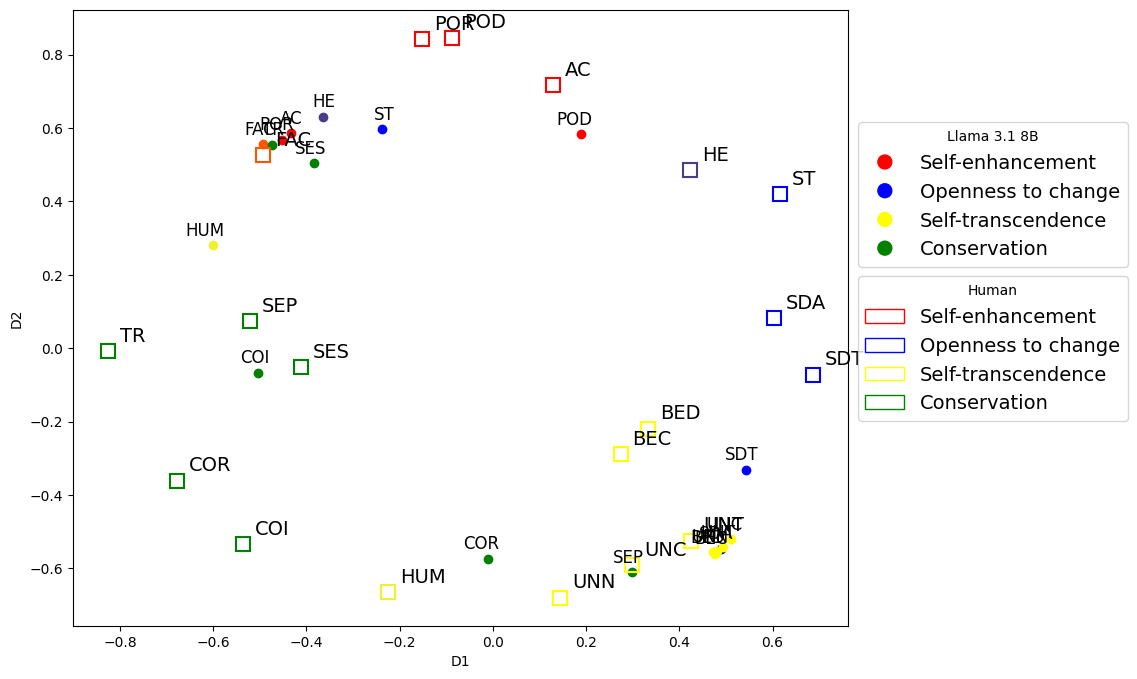}
            \caption{Names}
            \label{fig:Llama_names00}
            \captionsetup[sub]{labelformat=empty} 
        \end{subfigure}
    \end{minipage}%
    \begin{minipage}{0.5\linewidth}
        \centering
        \begin{subfigure}{\linewidth}
            \includegraphics[width=\linewidth]{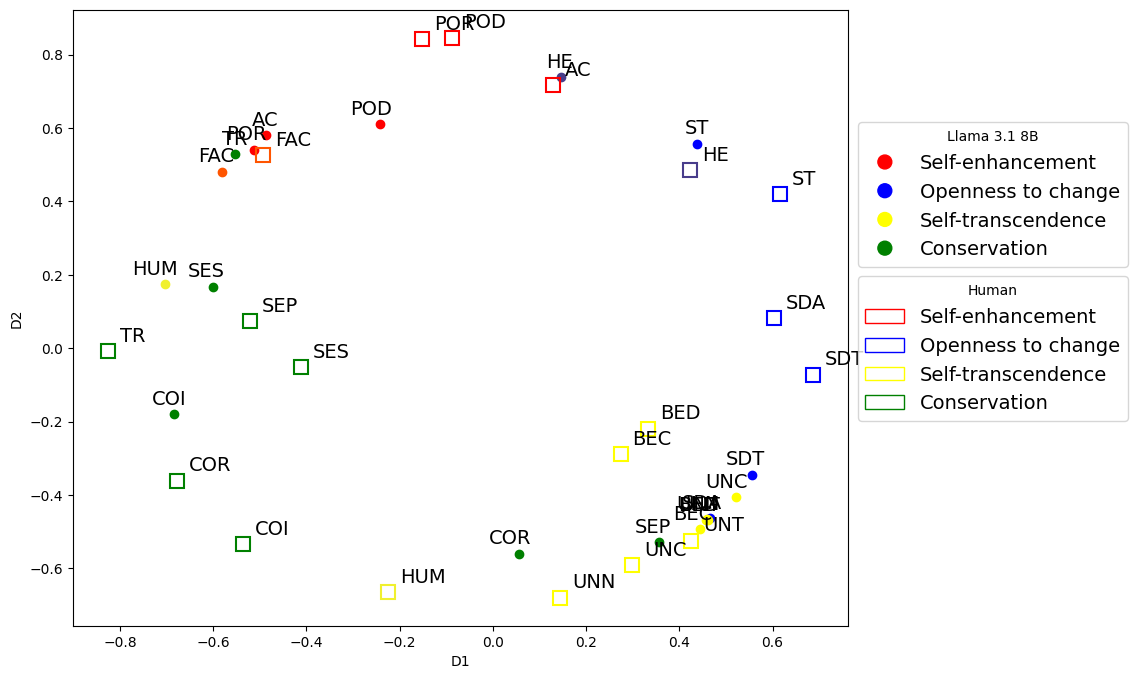}
            \caption{Demographic}
            \label{fig:Llama_demographic00}
        \end{subfigure}
        \begin{subfigure}{\linewidth}
            \includegraphics[width=\linewidth]{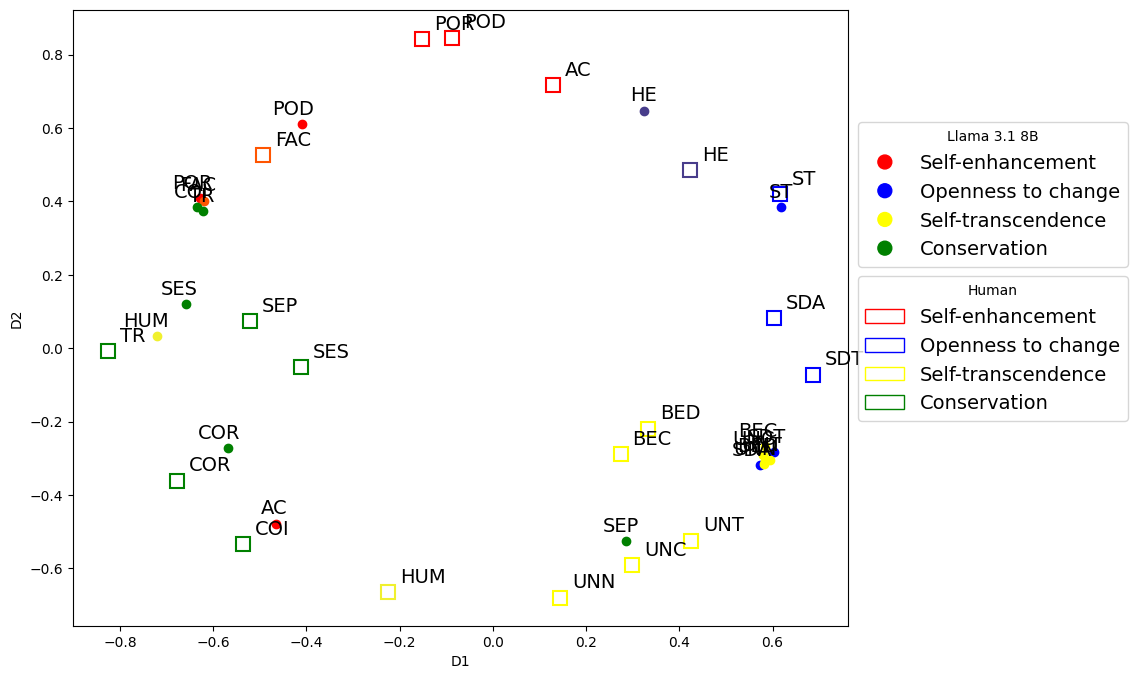}
            \caption{Generated Persona}
            \label{fig:Llama_persona00}
        \end{subfigure}
    \end{minipage}
    \caption{Comparison of the MDS results between human data \citep{schwartz2022measuring} and Llama 3.1 8B for all prompts, in the temperature $0.0$ condition.}
    \label{fig:llama_all_prompts}
\end{figure}

\begin{figure}[ht]
    \centering
    \begin{minipage}{0.5\linewidth}
        \centering
        \begin{subfigure}{\linewidth}
            \includegraphics[width=\linewidth]{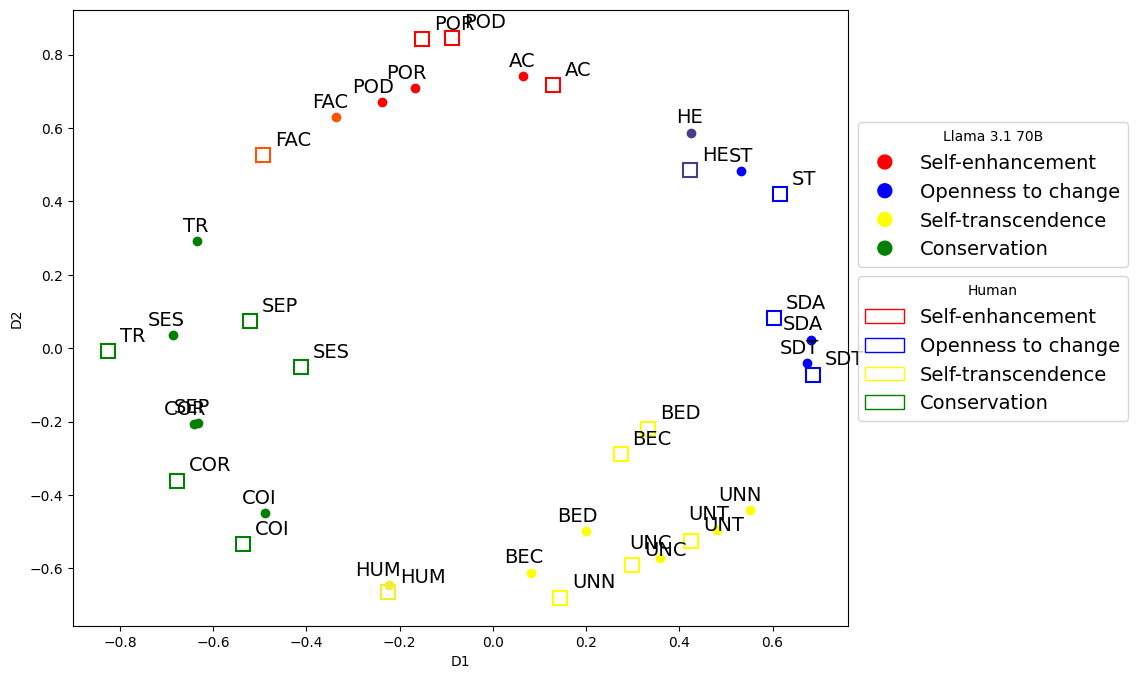}
            \caption{Value Anchor}
            \label{fig:Llama70b_bwvr00}
        \end{subfigure}
        \begin{subfigure}{\linewidth}
            \includegraphics[width=\linewidth]{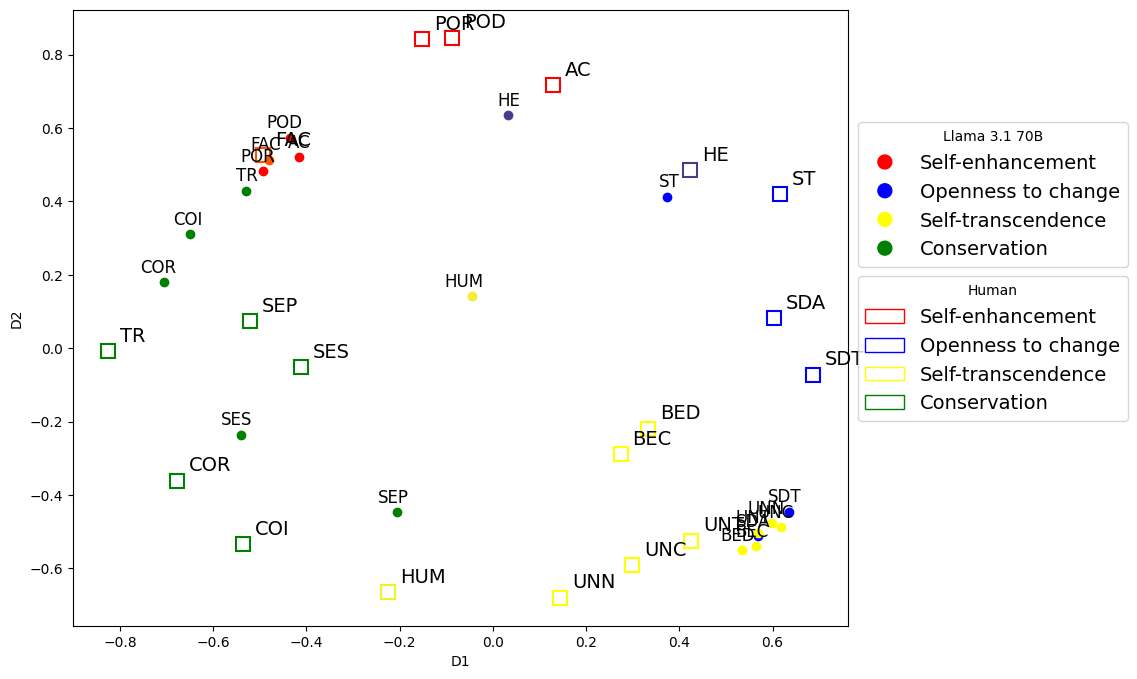}
            \caption{Names}
            \label{fig:Llama70b_names00}
            \captionsetup[sub]{labelformat=empty} 
        \end{subfigure}
    \end{minipage}%
    \begin{minipage}{0.5\linewidth}
        \centering
        \begin{subfigure}{\linewidth}
            \includegraphics[width=\linewidth]{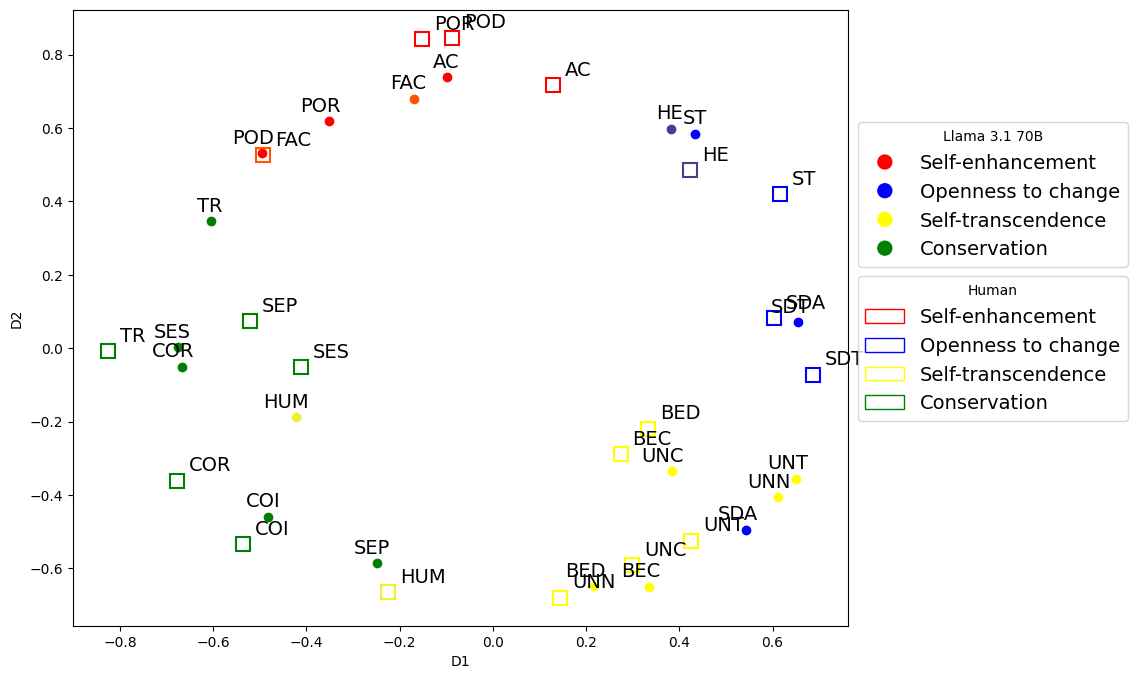}
            \caption{Demographic}
            \label{fig:Llama70b_demographic00}
        \end{subfigure}
        \begin{subfigure}{\linewidth}
            \includegraphics[width=\linewidth]{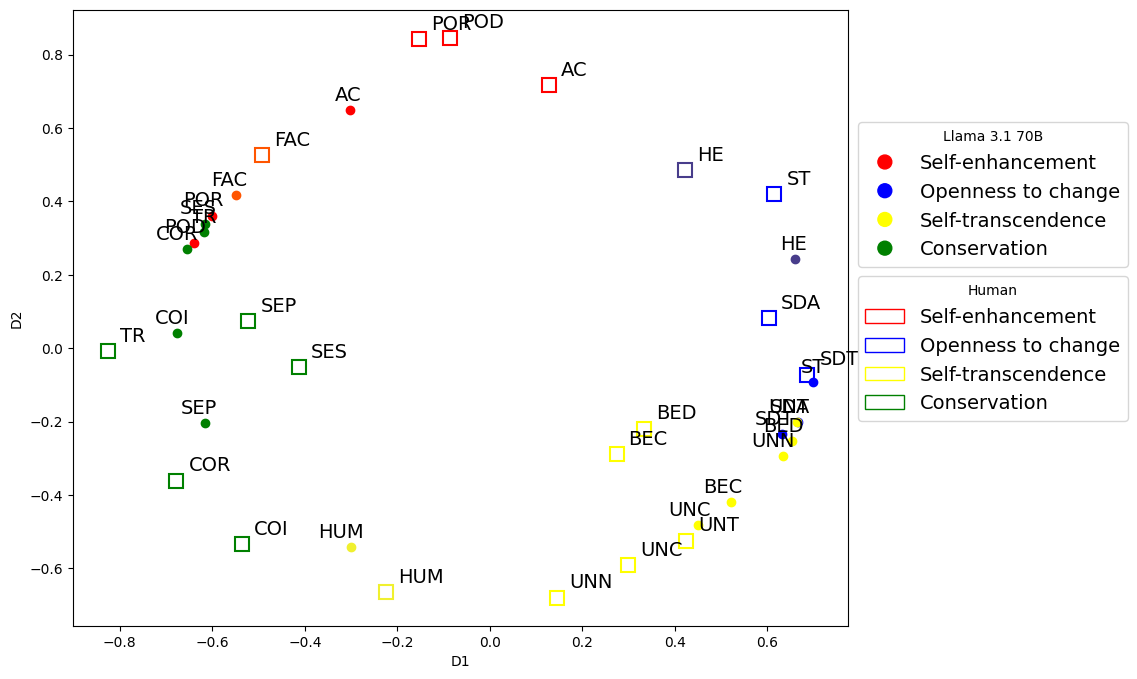}
            \caption{Generated Persona}
            \label{fig:Llama70b_persona00}
        \end{subfigure}
    \end{minipage}
    \caption{Comparison of the MDS results between human data \citep{schwartz2022measuring} and Llama 3.1 70B for all prompts, in the temperature $0.0$ condition.}
    \label{fig:llama70b_all_prompts}
\end{figure}

\begin{figure}[ht]
    \centering
    \begin{minipage}{0.5\linewidth}
        \centering
        \begin{subfigure}{\linewidth}
            \includegraphics[width=\linewidth]{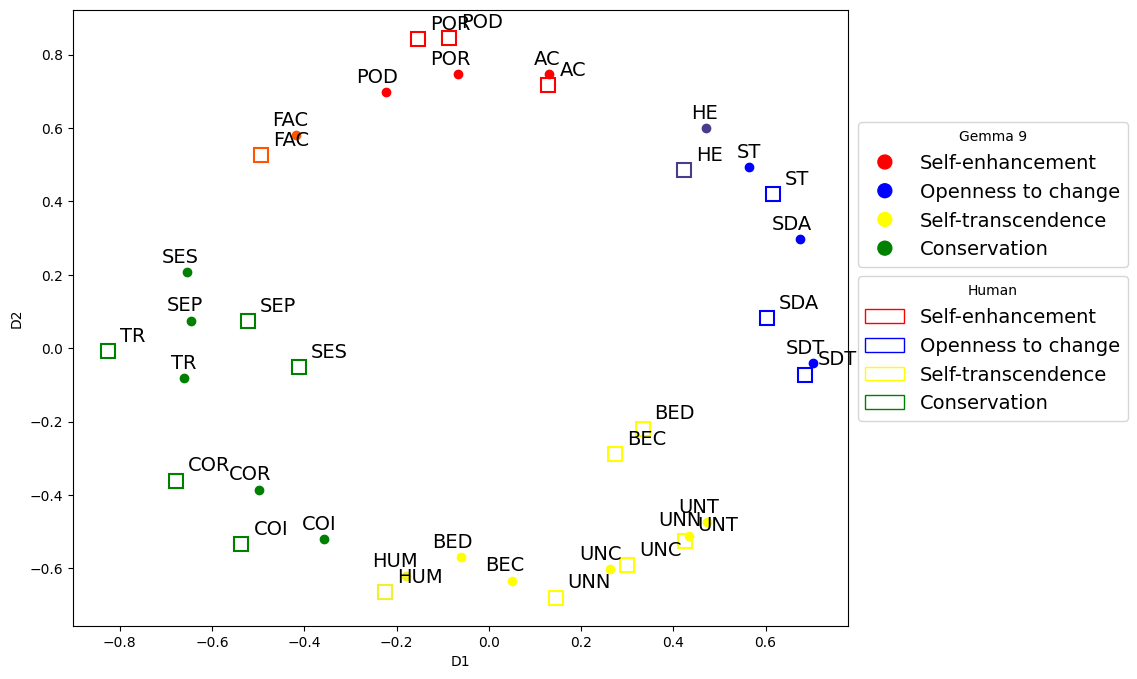}
            \caption{Value Anchor}
            \label{fig:Gemma9_bwvr00}
        \end{subfigure}
        \begin{subfigure}{\linewidth}
            \includegraphics[width=\linewidth]{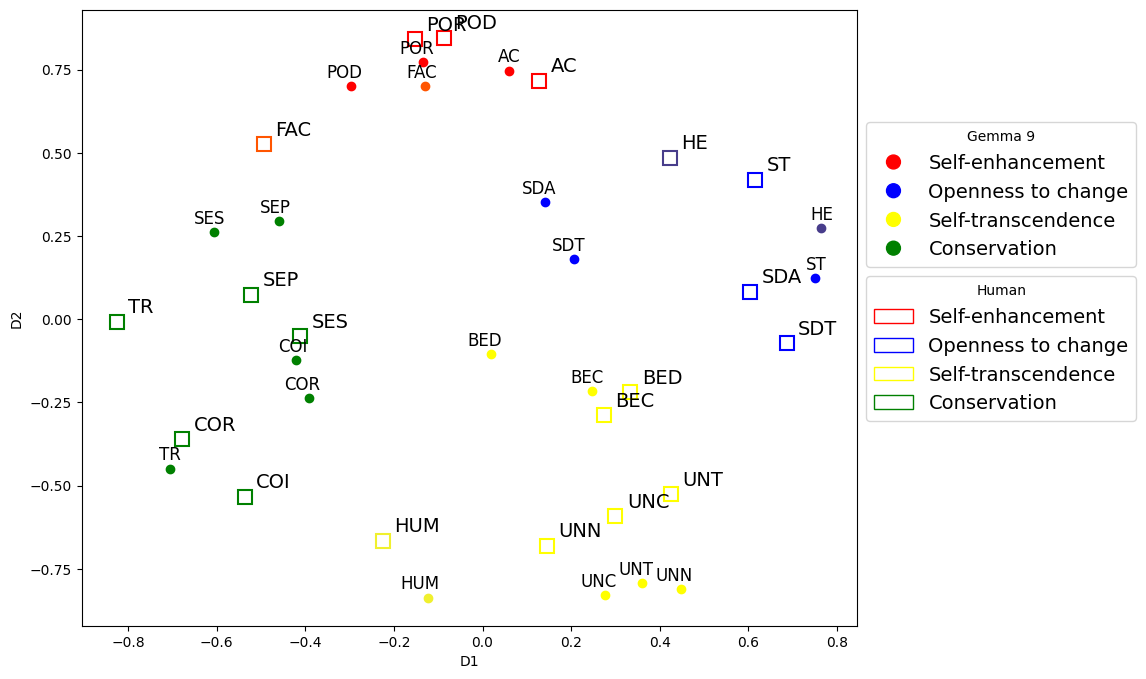}
            \caption{Names}
            \label{fig:Gemma9_names00}
            \captionsetup[sub]{labelformat=empty} 
        \end{subfigure}
    \end{minipage}%
    \begin{minipage}{0.5\linewidth}
        \centering
        \begin{subfigure}{\linewidth}
            \includegraphics[width=\linewidth]{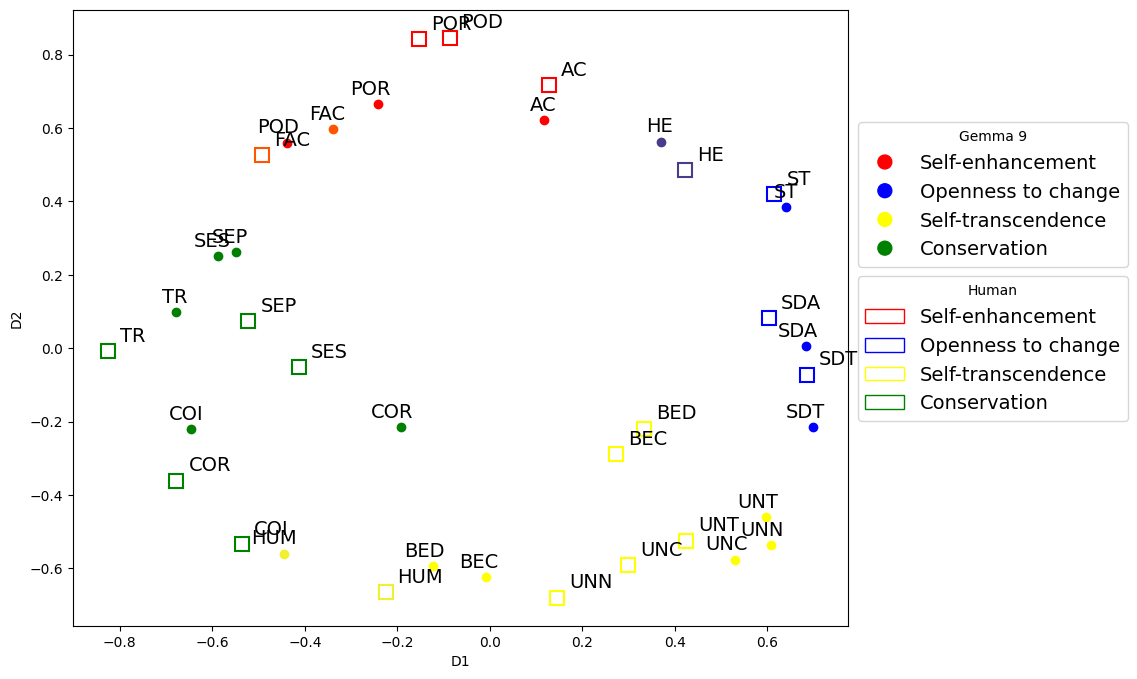}
            \caption{Demographic}
            \label{fig:Gemma9_demographic00}
        \end{subfigure}
        \begin{subfigure}{\linewidth}
            \includegraphics[width=\linewidth]{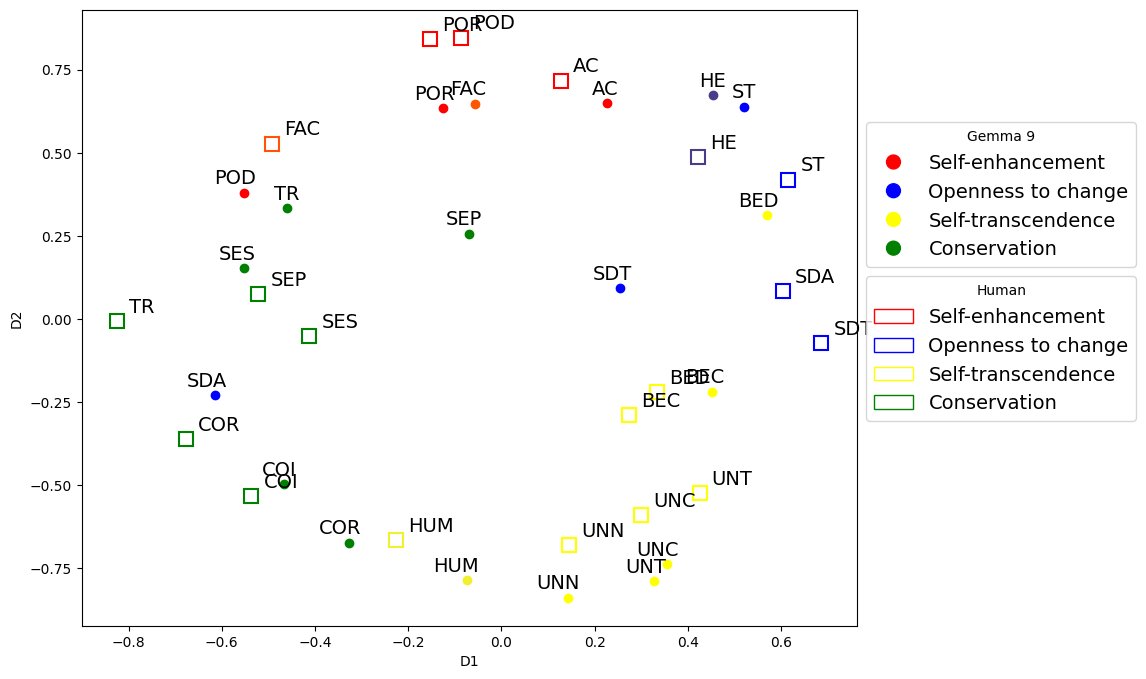}
            \caption{Generated Persona}
            \label{fig:Gemma9_persona00}
        \end{subfigure}
    \end{minipage}
    \caption{Comparison of the MDS results between human data \citep{schwartz2022measuring} and Gemma 2 9B for all prompts, in the temperature $0.0$ condition.}
    \label{fig:gemma9_all_prompts}
\end{figure}

\begin{figure}[ht]
    \centering
    \begin{minipage}{0.5\linewidth}
        \centering
        \begin{subfigure}{\linewidth}
            \includegraphics[width=\linewidth]{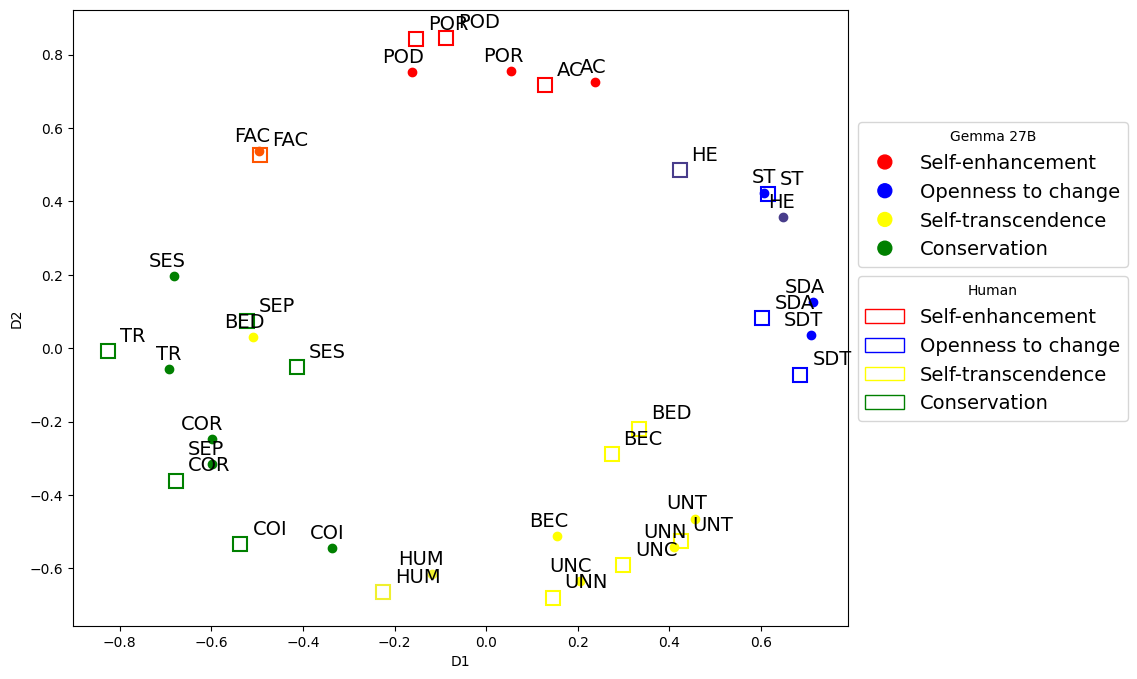}
            \caption{Value Anchor}
            \label{fig:Gemma27_bwvr00}
        \end{subfigure}
        \begin{subfigure}{\linewidth}
            \includegraphics[width=\linewidth]{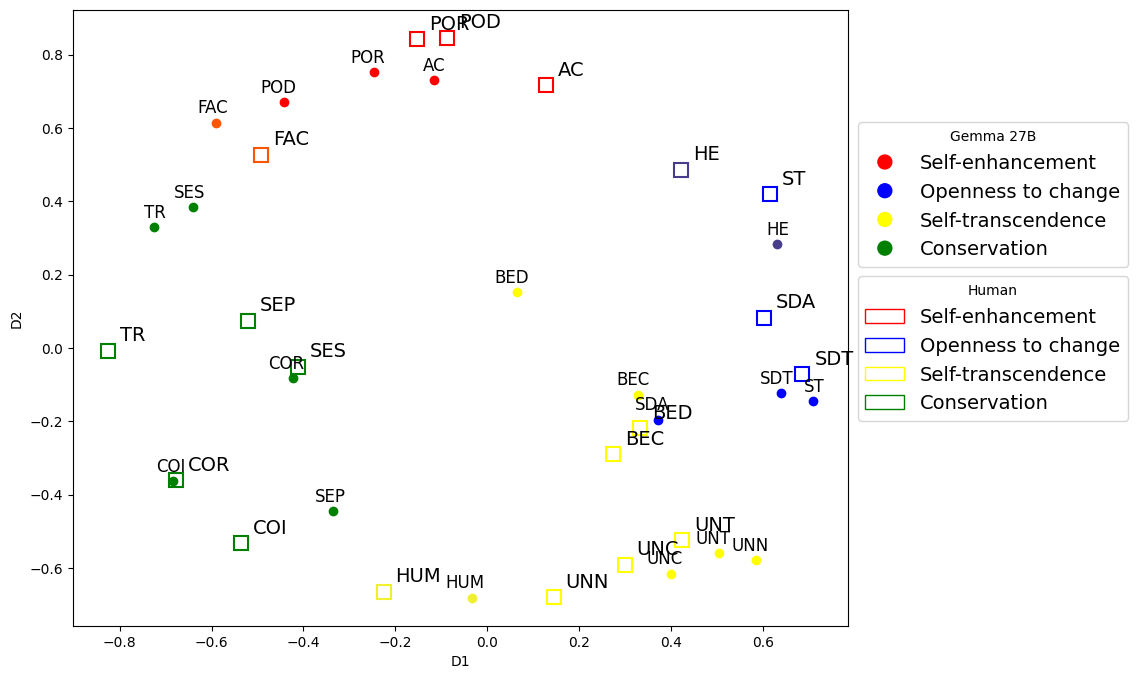}
            \caption{Names}
            \label{fig:Gemma27_names00}
            \captionsetup[sub]{labelformat=empty} 
        \end{subfigure}
    \end{minipage}%
    \begin{minipage}{0.5\linewidth}
        \centering
        \begin{subfigure}{\linewidth}
            \includegraphics[width=\linewidth]{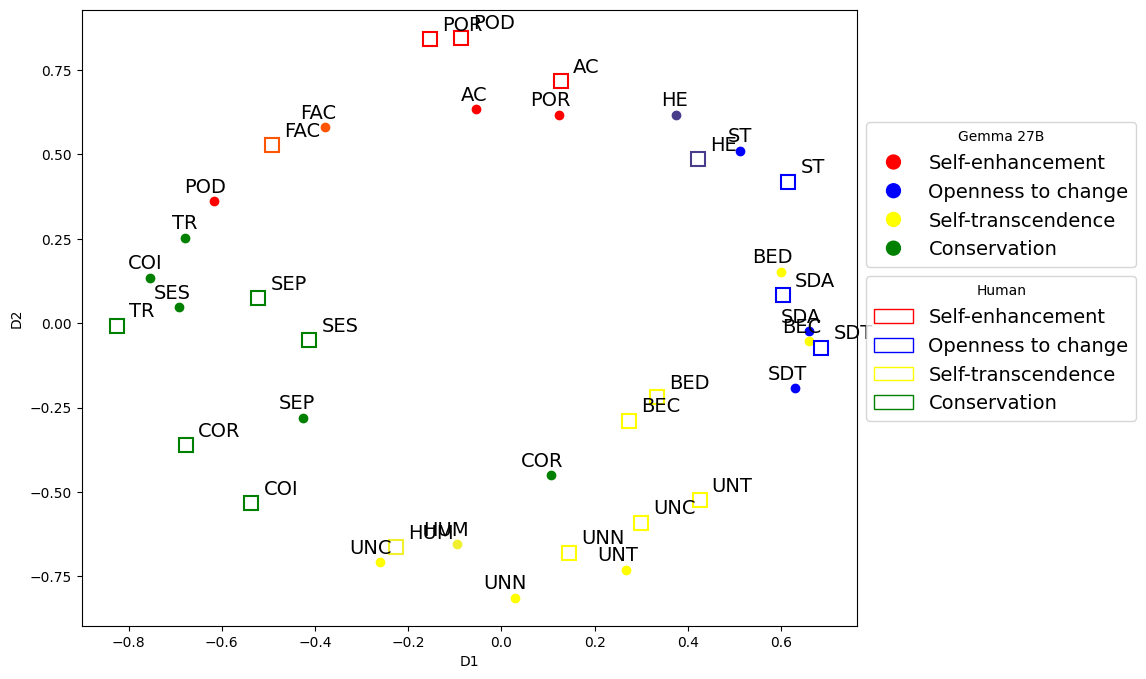}
            \caption{Generated Persona}
            \label{fig:Gemma27_persona00}
        \end{subfigure}
    \end{minipage}
    \caption{Comparison of the MDS results between human data \citep{schwartz2022measuring} and Gemma 2 27B for all prompts, in the temperature $0.0$ condition, with the exception of the Demographic prompt-(see Footnote 2).}
    \label{fig:gemma27_all_prompts}
\end{figure}

In the main text we provided the MDS plots for Gemini Pro for Value Anchor and Names. Here we provide further plots for Gemini Pro in \figref{fig:GeminiPro_demographic_persona}, all the GPT 4 plots in \figref{fig:GPT4_all_prompts}, all of the Llama 3.1 8B plots in \figref{fig:llama_all_prompts} and all of Llama 3.1 70B plots \figref{fig:llama70b_all_prompts}, and all of Gemma 2 9B plots \figref{fig:gemma9_all_prompts} and Gemma 2 27B plots \figref{fig:gemma27_all_prompts} for temperature $0.0$.

\section{\texorpdfstring{Comparing Batch and Sequential Prompting}{Comparing between one prompt vs serial prompting for Llama models}}
\begin{figure}[t]
    \centering
    \includegraphics[width=0.8\linewidth]{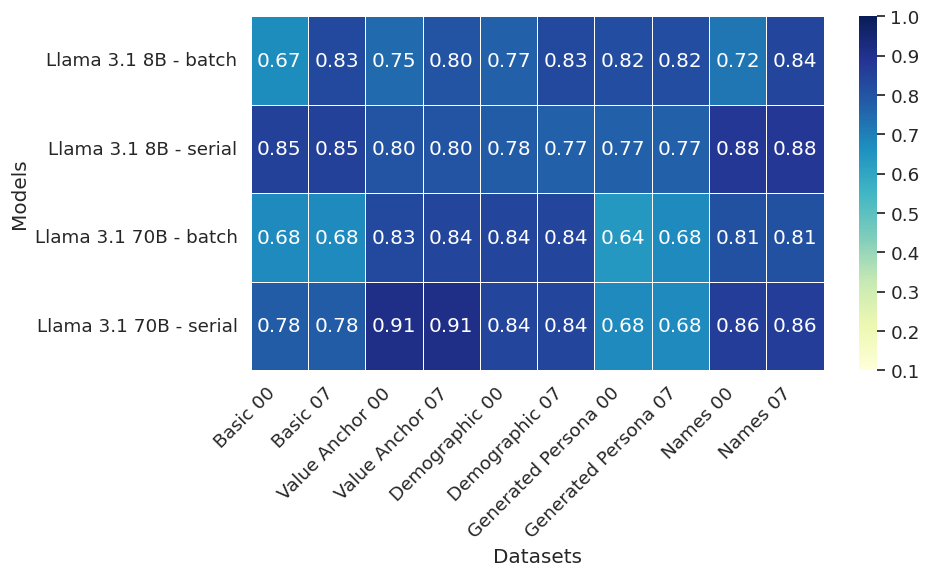}
    \caption{A heatmap of Spearman rank correlation between benchmark value hierarchies and dataset rankings for Llama 3.1 8B and 70B instruct for batch versus serial prompting methods, across temperature conditions. 
}    \label{fig:CorHeatmap_batchserial}
\end{figure}

In the main text, we focused exclusively on batch prompting, where all items from the questionnaire were presented in a single prompt. An alternative is to present the questions in sequence, and ask the model to answer a question as soon as it is presented. To investigate potential differences between batch and sequential prompting, we evaluated on Llama models (in commercial models, sequential prompting is more expensive than batch). The value-ranking results are summarized in  \figref{fig:CorHeatmap_batchserial}, and the value-correlation results in \tabref{tab:sum_square_scores_matrices_serial}.
Regarding the values rankings, significant differences were observed between two datasets in some instances (e.g., Llama 3.1 70B, Basic: \textit{z} = -2.64, \textit{p} \(=\) .004 ), while non-significant results were noted in other cases (e.g., Llama 3.1 8B, Value Anchor: \textit{z} = -0.83, \textit{p} \(=\) .203). This indicates that, overall, the ranking correlations are closely aligned, with no clear inclination toward either batch or sequential prompting as better replicating the human value hierarchy.
As for the value-correlations, it can be seen that the sequential prompts replicate the finding that the Value Anchor prompt best captures the circular structure of human values. Interestingly, for Llama 3.1 8B, batch prompting appeared to yield superior results. However, for Llama 3.1 70B, this was not the case across most prompts, suggesting that batch prompting may not consistently perform better across different models.

\comment{
\begin{figure}[t]
    \centering
    \includegraphics[width=0.8\linewidth]{cronbach's_alpha_00_llama_8_70_batch_serial.png}
    \caption{Comparing mean Cronbach's \(\alpha\) scores and standard deviations for Llama 3.1 8B and 70B instruct models: batch vs. serial prompting methods at temperature $0.0$, across different categories with human data \citep{schwartz2022measuring}}    \label{fig:cronbach_alpha_batchserial}
\end{figure}
}

\begin{table}[t]
\centering
\begin{tabular}{@{}lllll@{}}
\toprule
\multicolumn{5}{c}{\bf{Llama 3.1 8B}} \\
\cmidrule(lr){1-5}
 & \multicolumn{1}{c}{Value Anchor} & \multicolumn{1}{c}{Demographic} & \multicolumn{1}{c}{Generated Persona} & \multicolumn{1}{c}{Names} \\
\midrule
\multicolumn{5}{l}{\bf{Batch prompting}} \\
00 & 0.18 & 0.47 & 0.58 & 0.60 \\
07 & 0.16 & 0.47 & 0.58 & 0.57 \\
\midrule
\multicolumn{5}{l}{\bf{Serial prompting}} \\
00 & 0.18 & 0.54 & 0.65 & 0.61 \\
07 & 0.18 & 0.54 & 0.65 & 0.37 \\
\midrule
\multicolumn{5}{c}{\bf{Llama 3.1 70B}} \\
\cmidrule(lr){1-5}
 & \multicolumn{1}{c}{Value Anchor} & \multicolumn{1}{c}{Demographic} & \multicolumn{1}{c}{Generated Persona} & \multicolumn{1}{c}{Names} \\
\midrule
\multicolumn{5}{l}{\bf{Batch prompting}} \\
00 & 0.10 & 0.29 & 0.37 & 0.45 \\
07 & 0.10 & 0.22 & 0.40 & 0.44 \\
\midrule
\multicolumn{5}{l}{\bf{Serial prompting}} \\
00 & 0.14 & 0.26 & 0.20 & 0.48 \\
07 & 0.14 & 0.26 & 0.20 & 0.69 \\
\bottomrule
\end{tabular}
\caption{Sum of squared difference for MDS embeddings of humans and LLM.}
\label{tab:sum_square_scores_matrices_serial}
\end{table}

\newpage

\end{document}